\documentclass[10pt]{article} 
\usepackage[preprint]{tmlr}
 \usepackage{graphicx}
 \usepackage{algorithm}
\usepackage{algpseudocode}
\usepackage{float}
\usepackage{capt-of}

\usepackage{amsmath,amsfonts,bm}

\def\eqref#1{equation~\ref{#1}}

\def\1{\bm{1}}

\DeclareMathAlphabet{\mathsfit}{\encodingdefault}{\sfdefault}{m}{sl}
\SetMathAlphabet{\mathsfit}{bold}{\encodingdefault}{\sfdefault}{bx}{n}

\usepackage{hyperref}
\usepackage{url}

\title{Modular Bayesian Foundation Models\\
Bayesian time series encoders for OOD detection, routing and forecasting}

\author{\name Mame Diarra Toure \email mame.toure@mail.mcgill.ca \\
      \addr Department of Mathematics and Statistics\\
      McGill University
      \AND
      \name Quang Hung Pham \email Pham.QuangHung2@hydroquebec.com\\
      \addr Hydro-Quebec
      \AND 
      \name Mohamed Ryad Zemouri \email Zemouri.MohamedRyad@hydroquebec.com\\
      \addr Hydro-Quebec\\
      \AND
      \name David Stephens \email david.stephens@mcgill.ca\\
      \addr Department of Mathematics and Statistics\\
      McGill University\\
      \AND
      \name Luc Vouligny \email luc.vouligny@hydroquebec.com\\
      \addr Hydro-Quebec\\
      \AND
       \name Martin Gagnon \email gagnon.martin11@hydroquebec.com\\
      \addr Hydro-Quebec\\}

\def\month{MM}  
\def\year{YYYY} 
\def\openreview{\url{https://openreview.net/forum?id=XXXX}} 

\begin{document}
\raggedbottom

\maketitle

\begin{abstract}
Industrial systems continuously generate time series that vary in sampling frequency, scale, dynamics, and operating regime. Consequently a single time-series foundation model may not be able to represent all regimes effectively and retraining a large model for every new dataset or regime shift is computationally expensive. Modular foundation models address this issue by reusing specialised encoders trained on diverse datasets and by allowing the expansion of capabilities through the addition of new experts. This approach requires compact experts to minimize storage cost and a proper selection mechanism of experts for unseen data. We propose a modular foundation model framework that combines end-to-end Bayesian encoder training with uncertainty based routing. Using Singular Bayesian Neural Networks (SBNNs), we build a collection of compact Bayesian encoders whose epistemic uncertainty guides label-free selection. At the selected ranks, our Bayesian collection uses 63.8\% less storage than its deterministic counterpart. We evaluate this framework on time-series data through distribution shift, encoder selection, and probabilistic forecasting. Across seven incoming datasets, uncertainty-based routing is competitive with distance-based routing in terms of forecast MSE, while reducing average ECE from 0.086 to 0.072 and average NLL from 0.907 to 0.836. It also increases PICP90 from 82.28\% to 89.03\%. In a separate model-family comparison, we evaluate the SBNN pipeline against full-rank Bayes by Backprop, deep ensembles, and MC Dropout. The SBNN pipeline achieves an average MSE of 0.390, comparable with 0.396 for deep ensembles, 0.389 for MC dropout, and 0.443 for the full-rank Bayesian model. It uses 63.9\% fewer parameters than MC dropout, 81.9\% fewer than the full-rank model, and 92.8\% fewer than the deep ensemble. We also find that uncertainty-based routing helps shortlist useful
experts for the downstream task. However, the reconstruction ranking
within this shortlist does not necessarily carry over to forecasting.
\end{abstract}

\section{Introduction}
\label{sec:introduction}

Foundation models learn representations that are reusable across multiple datasets and tasks. For time series, TimesFM, Chronos, and Moirai provide pretrained
forecasters that transfer across datasets
\citep{Das2024TimesFM,Ansari2024Chronos,Woo2024Moirai},
while MOMENT extends representation reuse to tasks beyond
forecasting~\citep{Goswami2024}. However, time-series data can vary significantly in terms of sampling frequency, dynamics, or scale. Model rankings also vary across time-series domains, so strong aggregate benchmark performance need not identify the best model for a given application~\citep{syed2026position}. Modular foundation models allow for the separate training of specialized encoders, which can subsequently be reused in combination. This enables the collection to expand without the need to retrain existing encoders~\citep{Pfeiffer2023Modular}. However, the usefulness of an expanding collection depends on two questions: which experts should be used for unfamiliar data and how much does each expert cost to store? Selection is consequential: an encoder trained successfully on its source dataset need not provide a useful representation of a new one. Storing every encoder also incurs a cost that grows with the collection, even when only a small subset is deployed for a task.

In our framework, each expert is a Bayesian encoder, trained independently on a source dataset. Our contribution focuses on the experts themselves and the signal used to select them: our encoders carry weight posteriors and their epistemic uncertainty determines both their selection and their aggregation weights. The forecasting head is also Bayesian.
Each encoder stores its uncertainty on its source domain as a reference.  Thus, the selection and weighting of the frozen encoders requires no target labels or router training. We build on Singular Bayesian Neural Networks (SBNNs)~\citep{Toure2026}. For a layer weight matrix $W$ with $m$ rows and $n$ columns,
SBNNs use the factorization $W=AB^\top$.
Here, $A\in\mathbb{R}^{m\times r}$ and
$B\in\mathbb{R}^{n\times r}$ are Bayesian factor matrices,
and $r$ is the chosen rank. The variational parameter count scales as $\mathcal{O}(r(m+n))$ rather than $\mathcal{O}(mn)$. For $r<\min(m,n)$, the induced weight posterior is singular with respect to Lebesgue measure and introduces dependencies between weights even when the factor posterior is mean-field. Shared factors propagate uncertainty across connected weights.
Empirically, SBNNs improve out-of-distribution (OOD) detection
over full-rank Bayes by Backprop (BBB) across the evaluated
architectures.
In time-series forecasting, they also achieve better interval
coverage and calibration than deep ensembles and Rank-1 models,
while using substantially fewer parameters~\citep{Toure2026}. We inherit this parameterization and posterior construction from SBNNs. Our contribution is to adapt them to a modular collection of encoders and establish how their uncertainty serves shift assessment, expert selection, and downstream prediction.

Our framework relates to work on modular model reuse and expert
selection~\citep{Pfeiffer2023Modular,Pham2026Modular}.
Appendix~\ref{sec:related_work} discusses these connections,
Bayesian compression, uncertainty-based routing and downstream
transfer in detail.

\paragraph{Contributions.}
We develop a modular framework in which compact Bayesian encoders are trained end to end, retained as reusable experts, and combined with a Bayesian downstream head (\S\ref{sec:framework}). SBNN supplies the low-rank Bayesian parameterization; the present work studies how such experts can be selected and reused on incoming datasets. Our SBNN collection uses $63.8\%$\footnote{This counts encoders only. Including the forecasting head gives a reduction of 63.9\%.} fewer stored parameters than its deterministic counterpart with both posterior means and scales included (\S\ref{subsec:metrics_parameters}).

The Bayesian router uses each encoder's source-normalized epistemic
uncertainty to select the top-$K$ experts and assign their aggregation
weights (\S\ref{subsec:uncertainty_routing}), where $K$ is the number
of selected encoders. These decisions require no target labels,
router training, or encoder retraining. Across seven incoming datasets,
uncertainty-based routing remains competitive with source-distance
routing in forecasting mean squared error (MSE), while improving
expected calibration error (ECE)~\citep{Kuleshov2018Calibration},
negative log-likelihood (NLL)~\citep{GneitingRaftery2007}, and prediction
interval coverage probability at the nominal $90\%$ level (PICP@90).
Coverage increases from $82.28\%$ to $89.03\%$
(\S\ref{subsec:routing_comparison}).

In a separate model-family comparison, we evaluate the SBNN pipeline against a five-member ensemble~\citep{Lakshminarayanan2017},
MC Dropout~\citep{Gal2016}, and full-rank mean-field
Bayes by Backprop~\citep{Blundell2015} (\S\ref{sec:forecasting}).
SBNN maintains competitive forecasting accuracy and uncertainty
calibration with substantially fewer stored parameters.
Finally, the policy comparisons
(\S\ref{subsec:reconstruction_forecasting}) show that uncertainty helps identify a useful subset of experts for forecasting. Within this subset, lower reconstruction error does not consistently imply better forecasting performance.

\section{An Uncertainty-Aware Modular Bayesian Framework}
\label{sec:framework}

\subsection{Problem formulation and pipeline overview}
\label{subsec:problem_formulation}
We consider a collection of $M$ encoders pretrained independently
on different source datasets.
Each encoder $f_m(\cdot;\theta_m)$ maps an
input time-series window to a representation and has a variational
posterior $q_m(\theta_m)$ over its weights.
Here, $m\in\{1,\ldots,M\}$ indexes the encoders, and
$\theta_m$ denotes the weight parameters of encoder $m$. Several encoders may
share a source dataset while differing in architecture or training
configuration. Their learned representations do not necessarily have the same dimension.
Source pretraining uses self-supervised learning (SSL): training targets are constructed
from the inputs themselves. The pretext objective combines masked
reconstruction ~\citep{Vincent2008Denoising}, learned with a decoder, and a contrastive term
that compares representations of different views of the inputs~\citep{Chen2020SimCLR}.
Following \citet{Toure2026}, we learn Gaussian means and scales
for the low-rank factors, with factor-space Kullback--Leibler (KL)
regularization. For a layer with an $a\times b$ weight matrix and
rank $r$, the factors require $2r(a+b)$ stored parameters rather
than $ab$; biases and other parameters are counted separately.
We select compact rank configurations that remain competitive in
reconstruction accuracy and calibration
(Appendix~\ref{app:rank_kl_sweep}).
After pretraining, we retain the encoder posterior and compute
its source-domain uncertainty reference for routing.
Appendix~\ref{app:source_pretext_protocol} gives the loss definitions,
Bayesian parameterization, initialization and training settings.

Given an incoming dataset, our goal is to select and combine
encoders for downstream forecasting. We distinguish the unlabeled
input windows $\mathcal{X}_{\mathrm{route}}$ used for selection
from the supervised training pairs
$\mathcal{D}_{\mathrm{train}}=\{(x_i,y_i)\}_{i=1}^{N}$ used to fit the forecasting head.
Here, $N$ is the number of training pairs, and $y_i$ contains
the future observations to predict from input window $x_i$.
Routing uses unlabeled input windows, which may come from $\mathcal{D}_{\mathrm{train}}$. Expert selection is therefore label-free;
downstream head training is supervised.

The routing step returns a subset
$\mathcal{S}\subseteq\{1,\ldots,M\}$ of $K$ encoders and
nonnegative aggregation weights $\{\alpha_m\}_{m\in\mathcal{S}}$,
with $\sum_{m\in\mathcal{S}}\alpha_m=1$. Under our main policy,
selection and weighting use each encoder's epistemic uncertainty
on the incoming inputs relative to a stored source-domain reference.
The reference is computed before downstream reuse.
Routing requires no target labels, router training, or access
to the original source examples.

\begin{figure}[tbp]
\centering
\includegraphics[width=0.95\linewidth]{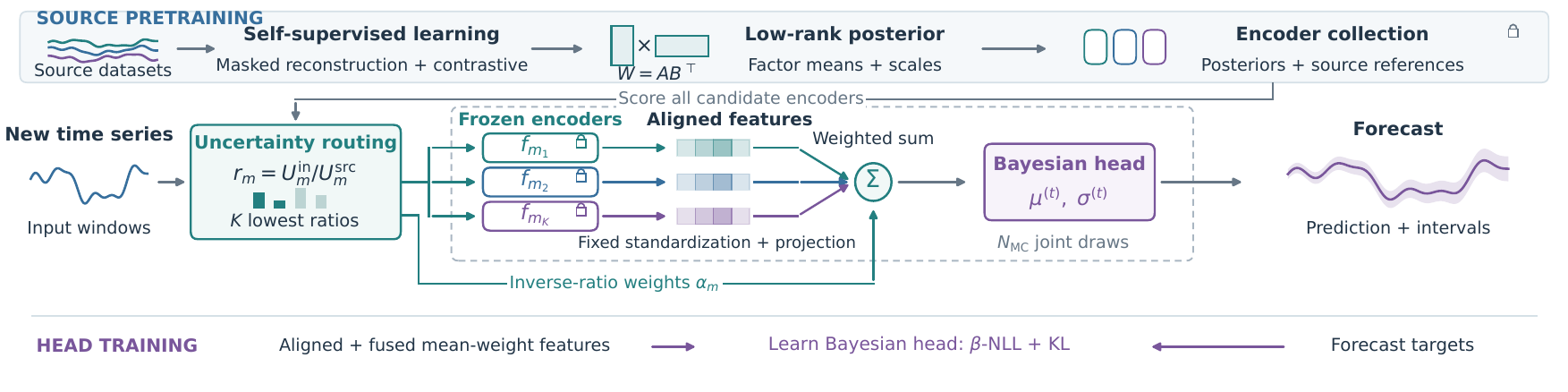}
\caption{\textbf{The modular Bayesian pipeline.}
Padlocks mark frozen posteriors whose weights are sampled.
The dashed boundary marks $N_{\mathrm{MC}}$ joint encoder--head draws;
selection, projections and aggregation weights remain fixed.}
\label{fig:modular_bayesian_pipeline}
\end{figure}

The selected representations are projected into a common space,
combined using the routing weights, and passed to a Bayesian
forecasting head. During head training, the encoder posteriors,
selected subset, and aggregation weights remain fixed.
At prediction, each draw samples the weights of all selected
encoders and the head. The resulting forecasts are combined
to estimate the predictive mean and uncertainty.
The reconstruction decoders are excluded from forecasting;
they are used only for pretraining and reconstruction evaluation.
Figure~\ref{fig:modular_bayesian_pipeline} summarizes the pipeline.

\subsection{Training-free routing with epistemic uncertainty}
\label{subsec:uncertainty_routing}

\begin{figure}[tbp]
\centering
\includegraphics[width=0.90\linewidth]{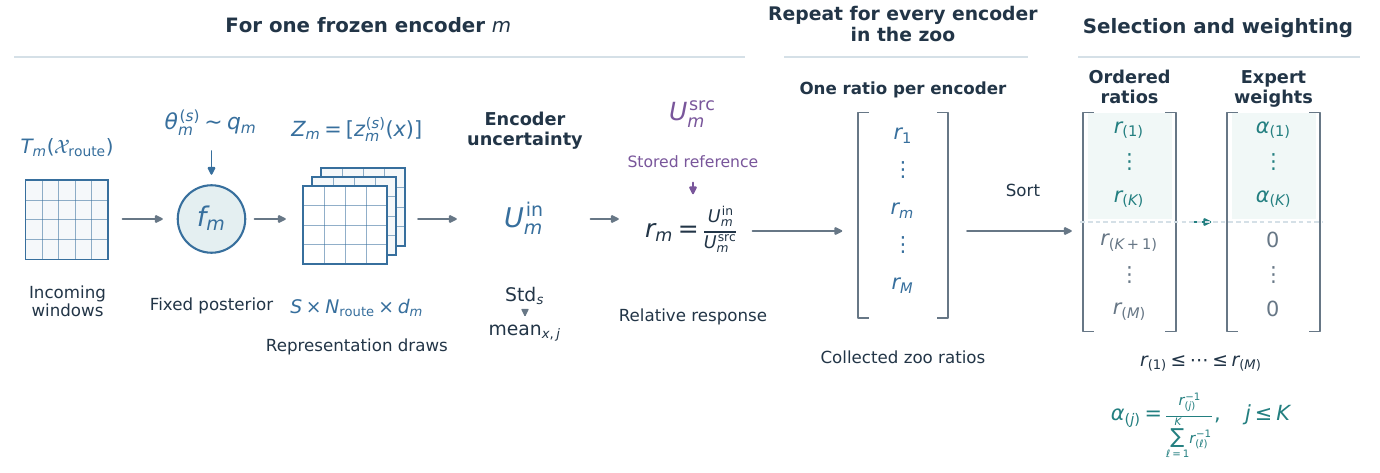}
\caption{\textbf{Training-free uncertainty routing.}
Posterior draws estimate uncertainty for each of the $M$ encoders.
The $K$ lowest source-normalized ratios receive inverse-ratio weights;
others receive zero weight. Subscript $(j)$ denotes rank $j$;
grids show array dimensions. Routing uses no target labels, decoder or head.}
\label{fig:uncertainty_routing}
\end{figure}

For an incoming dataset, we select experts by comparing each encoder's
uncertainty on the new inputs with its uncertainty on its own source
data. The resulting ratios determine both the selected set and its
aggregation weights. Figure~\ref{fig:uncertainty_routing} shows this
procedure.

For each encoder $m$, we draw $S$ weight samples from its frozen
variational posterior. The representation of an input window $x$ is
\begin{equation}
    z_m^{(s)}(x)
    = f_m\!\left(T_m(x);\theta_m^{(s)}\right),
    \qquad
    \theta_m^{(s)}\sim q_m,
    \quad s=1,\ldots,S,
    \label{eq:routing_latent_samples}
\end{equation}
where $T_m$ applies the formatting and preprocessing required by
encoder $m$. The input stays fixed across draws, so changes in the
representation come from the sampled weights. We measure this
variation by the standard deviation of each representation coordinate,
then average over coordinates and the input windows
in a set $\mathcal{X}$:
\begin{equation}
    U_m(\mathcal{X})
    = \frac{1}{|\mathcal{X}|\,d_m}
      \sum_{x\in\mathcal{X}}
      \sum_{j=1}^{d_m}
      \widehat{\operatorname{Std}}_{s=1,\ldots,S}
      \left[z_{m,j}^{(s)}(x)\right],
    \label{eq:latent_epistemic_uncertainty}
\end{equation}
where $d_m$ is the representation dimension. This statistic measures
encoder epistemic uncertainty. We compute it directly from the encoder representations.

Raw uncertainty levels differ across encoders. We therefore compare
each encoder with its own source reference, $U_m^{\mathrm{src}}$,
computed using the same statistic before downstream reuse.
For incoming windows $\mathcal{X}_{\mathrm{route}}$, the ratio is
\begin{equation}
    r_m
    = \frac{U_m(\mathcal{X}_{\mathrm{route}})}
           {U_m^{\mathrm{src}}}.
    \label{eq:source_normalized_uncertainty}
\end{equation}
A ratio above one indicates greater uncertainty than on the source
data; a ratio below one indicates less uncertainty. Only a scalar source uncertainty reference per encoder is needed
at routing time.

Our main policy selects the $K$ encoders with the smallest ratios.
For the selected set $\mathcal{S}$, normalized inverse-ratio weights are
\begin{equation}
    \alpha_m
    = \begin{cases}
        \displaystyle
        \frac{r_m^{-1}}{\sum_{j\in\mathcal{S}}r_j^{-1}},
        & m\in\mathcal{S},\\[8pt]
        0, & m\notin\mathcal{S}.
      \end{cases}
    \label{eq:inverse_ratio_routing_weights}
\end{equation}
Lower ratios receive larger weights. These expressions assume positive
source references and ratios. The choices of $K$ and the comparison
policies are given in \S\ref{subsec:baselines_protocol}.

Routing is performed once per incoming dataset. It requires posterior
forward passes through every candidate encoder, without parameter
updates or target labels. Selection uses the uncertainty ratios. Reconstruction errors and
forecasting targets are used for evaluation. The selected set and its weights
then remain fixed during downstream training and prediction.

\paragraph{Remark (family comparison).}
Each uncertainty family routes independently; the deterministic
pipeline reuses SBNN selections and weights
(\S\ref{subsec:baselines_protocol}).
Its heteroscedastic regression head predicts the target $y$
and an input-dependent standard deviation $\sigma$.

\subsection{Representation aggregation and downstream prediction}
\label{subsec:aggregation_prediction}

The selected encoders may produce representations of different
dimensions. For each encoder, we compute a coordinate-wise mean
and a single global standard deviation from its posterior-mean
representations on the target training split.
We subtract the mean and divide by the standard deviation.

We then project each representation into 64 dimensions using
a fixed random matrix. Its entries are drawn independently
from a standard Gaussian distribution and divided by the square
root of the encoder's output dimension.
An encoder-specific seed fixes this matrix.
These data-independent projections give the head a fixed 64-dimensional input (Appendix~\ref{app:projection_fusion}).
The normalization and projection remain fixed during head
training and test-time sampling.

Let $P_m$ denote this standardization and projection for encoder $m$.
For the selected encoder weights
$\theta_{\mathcal{S}}=\{\theta_m:m\in\mathcal{S}\}$,
the aggregated representation is
\begin{equation}
    h(x;\theta_{\mathcal{S}})
    =
    \sum_{m\in\mathcal{S}}
    \alpha_m P_m\!\left(f_m(T_m(x);\theta_m)\right).
    \label{eq:representation_aggregation}
\end{equation}
The reconstruction decoders are not used in this stage.

During head training, the encoders provide fixed representations
computed with their posterior-mean weights. A Bayesian forecasting head, $g$, learns from these aggregated representations and the target values in $\mathcal{D}_{\mathrm{train}}$.
Each training window produces one aggregated representation,
paired with its own forecast target.
The head's weight parameters $\omega$ have variational posterior
$q_\phi(\omega)$, where $\phi$ collects the learned posterior
means and scales.
The encoder posteriors, projections, and routing weights remain
fixed. The lower strip of Figure~\ref{fig:modular_bayesian_pipeline}
shows this training stage.
The final head has two outputs for each forecast
coordinate: a predictive mean $\mu$ and a positive standard deviation
$\sigma$. This second output represents aleatoric uncertainty.
We train both outputs with $\beta$-NLL
\citep{Seitzer2022PitfallsOfUncertainty}, using $\beta=0.5$.
For the Bayesian head, the training objective also includes
factor-space KL regularization, as in
\S\ref{subsec:bayesian_encoder_learning}.
Appendix~\ref{app:forecasting_loss_transition} gives the full
objective and the motivation for replacing the initial MSE-trained head.

At test time, we sample the entire pipeline inside the dashed boundary
in Figure~\ref{fig:modular_bayesian_pipeline}.
For each draw \(t\), we sample encoder weights
\(\theta_m^{(t)}\sim q_m\), \(m\in\mathcal{S}\), and head weights
\(\omega^{(t)}\sim q_\phi\). We compute \(h^{(t)}(x)\) using
Equation~\ref{eq:representation_aggregation} with these sampled encoder weights.
The head returns
\(\left(\mu^{(t)}(x),\sigma^{(t)}(x)\right)
=g\!\left(h^{(t)}(x);\omega^{(t)}\right)\).
We repeat this procedure for \(N_{\mathrm{MC}}\) Monte Carlo draws.
The selected experts and their aggregation weights do not change
between draws. The point forecast is the mean of the sampled
predictions,
\(\widehat{\mu}(x)=N_{\mathrm{MC}}^{-1}
\sum_{t=1}^{N_{\mathrm{MC}}}\mu^{(t)}(x)\).

Variation in \(\mu^{(t)}(x)\) estimates the epistemic uncertainty
propagated through the encoders and the head:
\begin{equation}
    \widehat{v}_{\mathrm{epi}}(x)=
    \frac{1}{N_{\mathrm{MC}}}
    \sum_{t=1}^{N_{\mathrm{MC}}}
    \left(\mu^{(t)}(x)-\widehat{\mu}(x)\right)^2.
    \label{eq:predictive_epistemic_variance}
\end{equation}
The aleatoric variance comes from the head's standard-deviation
output,
\(\widehat{v}_{\mathrm{alea}}(x)=N_{\mathrm{MC}}^{-1}
\sum_{t=1}^{N_{\mathrm{MC}}}\left(\sigma^{(t)}(x)\right)^2\).
These calculations are performed separately for each forecast
coordinate. The total predictive variance is
\begin{equation}
    \widehat{v}_{\mathrm{pred}}(x)=
    \widehat{v}_{\mathrm{epi}}(x)+\widehat{v}_{\mathrm{alea}}(x).
    \label{eq:total_predictive_variance}
\end{equation}

Algorithm~\ref{alg:joint_prediction} in
Appendix~\ref{app:family_forecasting_details} gives the complete
prediction procedure after routing and head training.

\section{Experimental Setup}
\label{sec:experimental_setup}

We evaluate encoder storage, source reconstruction, distribution-shift
response and downstream forecasting. Routing-policy comparisons use SBNN encoders throughout.

\subsection{Source domains, encoder collections, and incoming datasets}
\label{subsec:datasets_encoders}
We pretrain encoders on four source datasets: Acoustic,
HAR (human activity recognition), Torque45 and
ETT (electricity transformer temperature).
We chose these datasets to cover different measured quantities,
operating regimes, and temporal dynamics.
The SBNN collection contains 17 encoders
(Table~\ref{tab:source_encoder_inventory}).
Several experts share a source dataset but differ in architecture
or training configuration.

\begin{table}[!htbp]
\centering
\begin{minipage}[t]{0.38\linewidth}
\vspace{0pt}
\centering
\footnotesize
\caption{\footnotesize\textbf{SBNN encoder inventory by source.}}
\label{tab:source_encoder_inventory}
\setlength{\tabcolsep}{2pt}
\renewcommand{\arraystretch}{1.04}
\begin{tabular}{@{}p{0.19\linewidth}p{0.56\linewidth}r@{}}
\hline
Source & Signal & Encoders \\
\hline
Acoustic & Sound recordings & 7 \\
HAR & Human motion & 3 \\
Torque45 & Torque & 3 \\
ETT & Oil temperature & 4 \\
\hline
Total & & 17 \\
\hline
\end{tabular}
\end{minipage}\hfill
\begin{minipage}[t]{0.60\linewidth}
\vspace{0pt}
\centering
\footnotesize
\setlength{\tabcolsep}{4pt}
\renewcommand{\arraystretch}{1.12}
\caption{\footnotesize\textbf{Stored encoder parameters.}
Counts sum the 17 matched encoders per family, grouped by source.
Bayesian counts include posterior means and scales;
ensemble counts include all five members.
Decoders and heads are excluded.
Reductions are relative to deterministic encoders.
Bold marks the smallest count.}
\label{tab:encoder_storage}
\setlength{\tabcolsep}{2pt}
\renewcommand{\arraystretch}{1.04}
\begin{tabular}{@{}lrrrrr@{}}
\hline
Source & \shortstack{Deterministic /\\MC Dropout}
& SBNN & \shortstack{Full-rank\\BBB}
& \shortstack{Ensemble\\(5 members)} & Reduction \\
\hline
Acoustic & 3,064,256 & \textbf{1,048,002} & 6,124,098 & 15,321,280 & 65.8\% \\
HAR & 1,082,824 & \textbf{571,844} & 2,165,188 & 5,414,120 & 47.2\% \\
Torque45 & 555,289 & \textbf{60,088} & 1,110,586 & 2,776,445 & 89.2\% \\
ETT & 116,368 & \textbf{63,788} & 232,736 & 581,840 & 45.2\% \\
\hline
Total & 4,818,737 & \textbf{1,743,722} & 9,632,608 & 24,093,685 & 63.8\% \\
\hline
\end{tabular}
\end{minipage}
\end{table}

Incoming datasets were not used during source pretraining.
The seven incoming datasets are Torque10, Torque90, WTH (weather),
ETTh2, ECL (electricity consumption), vibration, and EMG
(electromyographic signals). Torque10 and Torque90 represent
operating conditions different from Torque45; ETTh2 belongs to
the ETT dataset family.
Acoustic, Torque45, Torque10, Torque90 and vibration were recorded
by an industrial partner; the remaining datasets are public.
Appendix~\ref{app:dataset_protocol} gives dataset descriptions,
origins and references.
Storage comparisons use the same source and architecture composition
across families, following \S\ref{subsec:metrics_parameters}.

\subsection{Baselines and evaluation protocol}
\label{subsec:baselines_protocol}

\paragraph{Model families.}
We compare SBNN with deterministic models, MC Dropout, full-rank
Bayes by Backprop (BBB), and five-member deep ensembles.
Full-rank refers to the weight parameterization; BBB uses a
mean-field Gaussian posterior over these weights.
For each arrival, SBNN, MC Dropout, BBB and deep ensembles
select their own $K$ experts and compute their own aggregation
weights. The deterministic pipeline reuses SBNN selections and weights. Each family supplies its corresponding encoders and
forecasting head, with the same head architecture across families
and the family's weight parameterization.
The comparison covers reconstruction, forecasting accuracy,
predictive uncertainty, and storage.
\paragraph{Source-distance score.}
We represent each dataset by a statistical vector
$\phi(\mathcal{X})$ containing its mean, standard deviation,
10th, 50th and 90th percentiles, and autocorrelations
at lags 1, 5 and 10.

For source $s$, let $\mathcal{X}_s$ denote its reference
windows. Let $c_j$ be the standard deviation of descriptor
coordinate $j$ across the four source datasets.
We compare the source and incoming descriptors using
\begin{equation}
    D_s(\mathcal{X})
    =
    \left[
        \frac{1}{8}
        \sum_{j=1}^{8}
        \left(
            \frac{
                \phi_j(\mathcal{X})-\phi_j(\mathcal{X}_s)
            }{c_j}
        \right)^2
    \right]^{1/2}.
    \label{eq:source_distance_score}
\end{equation}
The coordinate scales remain fixed across comparisons.
Lower distances indicate more similar dataset statistics,
and a dataset's distance to itself is zero.
Appendix~\ref{app:shift_protocol} details the descriptor
construction and scaling.
\paragraph{Routing policies.}
We compare uncertainty-based routing with a source-distance baseline
using SBNN encoders in both pipelines. Both select $K$ experts
from the same eligible pool, ranked by source-normalized epistemic
ratios (\S\ref{subsec:uncertainty_routing}) or statistical-feature
distance, with normalized inverse-score weights.
The distance baseline does not reproduce the full two-stage router
of \citet{Pham2026Modular}.
Appendix~\ref{app:routing_protocol_details} gives the
source-distance calculation and tie-handling rule.
The comparison changes selection and weighting together and
therefore evaluates each routing policy as a whole.

We also compare source-diverse selection, random selection, and
higher-uncertainty controls to examine whether a policy that supports
reconstruction also supports forecasting.
Random selection from the full candidate pool and from a
low-uncertainty shortlist are separate controls.
Table~\ref{tab:selection_policies} gives the selection and weighting
rules. Values of $K$, thresholds, and candidate pools are
specified with the corresponding results.

\paragraph{Head training and prediction.}
For the distance-versus-uncertainty comparison, both pipelines
use the same input windows, target splits, head architecture,
training settings, and seeds. Neither router uses forecast targets.
All final forecasting heads use $\beta$-NLL with $\beta=0.5$;
Bayesian heads also include KL regularization.
Forecasting is evaluated over three seeds: $40$, $41$, and $42$.
SBNN prediction uses $N_{\mathrm{MC}}=100$ joint encoder--head
draws, following \S\ref{subsec:aggregation_prediction}.
\subsection{Metrics and parameter accounting}
\label{subsec:metrics_parameters}

We evaluate reconstruction, forecasting accuracy and predictive
uncertainty separately. For reconstruction on incoming data, the
policy score is the weighted average of the selected encoders'
individual reconstruction errors,
$R_{\mathcal{S}}=\sum_{m\in\mathcal{S}}\alpha_m R_m$.
We compute each encoder's error before taking the weighted average.
Reconstruction comparisons are made within each incoming dataset.

Forecasting metrics use the standardized target scale.
We report MSE for point accuracy, expected calibration error
(ECE), predictive negative log-likelihood (NLL), prediction
interval coverage probability at 90\% (PICP@90), and mean
prediction interval width at 90\% (MPIW@90).
Lower MSE, ECE and NLL are preferred.
Coverage should approach the nominal 90\%; interval width is
interpreted alongside coverage.
Forecasting intervals and NLL use a Gaussian approximation
with the total predictive variance from
\S\ref{subsec:aggregation_prediction}.
The reported NLL is distinct from the $\beta$-NLL training loss.
For distribution shift, we report Spearman correlations between dataset distance and the uncertainty ratio.
Exact definitions and averaging conventions are given in
Appendix~\ref{app:evaluation_metrics}.

Parameter counts include frozen encoder weights, Bayesian
posterior means and scales, and all five ensemble members.
Complete forecasting-pipeline counts also include the head
and exclude reconstruction decoders and training state.
At equal numerical precision, parameter-count ratios equal
parameter-storage ratios.
Appendix~\ref{app:posterior_storage} gives the detailed accounting.

Forecasting results report means and sample standard deviations
over three seeds. Overall means give equal weight to the seven
incoming datasets; routing comparisons pair runs by dataset
and seed.

\section{Results}
\label{sec:results}

We evaluate encoder storage and the use of uncertainty for expert
selection. We first compare storage, reconstruction accuracy and
uncertainty calibration on the source datasets.
We then examine how encoder uncertainty responds to distribution
shift. The downstream experiments first compare forecasting quality
and parameter cost across model families, then compare
uncertainty-based routing with source-distance routing.
Finally, we compare selection and weighting policies under both
reconstruction and forecasting to examine whether their rankings
carry over between tasks.

\subsection{Storage and source-task performance of compact encoders}
\label{subsec:source_reconstruction}

Our starting point was an existing collection of deterministic encoders. We used SBNN to build compact Bayesian counterparts that reduce storage and provide uncertainty for expert selection. The comparison includes the
corresponding deterministic, MC Dropout, full-rank BBB and five-member
ensemble models on Acoustic, HAR, Torque45 and ETT.
Reconstruction uses each encoder's decoder. The decoder is
subsequently discarded from the forecasting pipeline.

\paragraph{Encoder storage.}
Table~\ref{tab:encoder_storage} counts the stored encoder parameters.
Bayesian counts include posterior means and scales, and ensemble
counts include all five members. Decoders, forecasting heads and
optimizer state are excluded.

At the selected ranks, the 17-encoder SBNN collection contains
1,743,722 parameters, compared with 4,818,737 for the matched
deterministic collection: a reduction of 63.8\%. At equal numerical precision,
parameter storage falls by the same proportion. MC Dropout stores
the same number of parameters as the deterministic model.
SBNN is smaller than every compared family on each source.

\begin{table}[!htbp]
\centering
\vspace{0pt}
\centering
\footnotesize
\renewcommand{\arraystretch}{1.15}
\caption{\footnotesize\textbf{Selection policies and weights.}
Each policy selects $K$ encoders. Top and bottom denote the lowest
and highest scores. The shortlist contains ratios
$r_m\leq\tau_{\mathrm{route}}$. Weights are normalized over selected encoders.}
\label{tab:selection_policies}
\setlength{\tabcolsep}{2pt}
\renewcommand{\arraystretch}{1.04}
\begin{tabular}{@{}p{0.28\linewidth}p{0.52\linewidth}p{0.16\linewidth}@{}}
\hline
Policy & Selection rule & Weights \\
\hline
Top-$K$ uncertainty
& $K$ lowest source-normalized uncertainty ratios
& Inverse-ratio \\
Top-$K$ distance
& $K$ lowest source-distance scores
& Inverse-distance \\
Source-diverse top-$K$
& Lowest-ratio encoder per source, then $K$ lowest ratios
& Inverse-ratio \\
Random-$K$, full pool
& Sample $K$ encoders without replacement
& Inverse-ratio \\
Random-$K$, shortlist
& Sample $K$ encoders without replacement
& Uniform \\
Bottom-$K$, shortlist
& $K$ highest uncertainty ratios within the shortlist
& Uniform \\
Bottom-$K$ uncertainty
& $K$ highest uncertainty ratios in the eligible pool
& Uniform \\
\hline
\end{tabular}
\end{table}

Figure~\ref{fig:source_encoder_storage} in
Appendix~\ref{app:posterior_storage} compares all five families.
A deterministic model with the same architecture and ranks would use fewer parameters than SBNN. SBNN also supplies the epistemic uncertainty used for expert selection, addressing both storage and routing. Relative to full-rank deterministic encoders, the reduction ranges
from 45.2\% to 89.2\%.
It depends on the architecture and the retained rank profile.
The rank controls the tradeoff between compression, reconstruction accuracy and calibration. The ETT rank sweep in
Appendix~\ref{app:rank_kl_sweep} shows how different rank
configurations affect these metrics.

\paragraph{Reconstruction accuracy and calibration.}
On source reconstruction, the selected SBNN encoders achieve
the lowest ECE and coverage closest to the nominal 90\% level
on three of the four source domains; HAR is the exception.
This comes with a source-dependent increase in reconstruction
error. SBNN therefore offers better uncertainty calibration at a lower storage cost.
Low reconstruction error can coexist with overconfident uncertainty
estimates, so we prioritize uncertainty quality alongside accuracy.
Appendix~\ref{app:source_results} gives the complete source comparison
and reconstruction plots.

\subsection{Epistemic response to distribution shift}
\label{subsec:distribution_shift}

Source-task calibration evaluates uncertainty on familiar data.
Figure~\ref{fig:encoder_shift_response} compares statistical-feature
distance (Equation~\ref{eq:source_distance_score}) with the
source-normalized epistemic uncertainty of encoders trained on
Torque45, Acoustic and ETT across eight domains.
Uncertainty increases with distance overall, with local reversals
in the ordering. Acoustic saturates faster.
No decoder or forecasting head is involved.
Appendix~\ref{app:shift_protocol} gives the protocol, correlations,
and per-domain values.

Shift assessment and routing make different comparisons.
Here, we vary the incoming dataset while holding the encoder fixed.
Routing compares several encoders on the same incoming dataset.
Each encoder has its own learned response to unfamiliar inputs.

\subsection{End-to-end forecasting across model families}
\label{subsec:forecasting}
\label{sec:forecasting}

We now use the frozen encoders to forecast incoming time series.
The comparison uses top-4 uncertainty routing
(Equation~\ref{eq:source_normalized_uncertainty}) and inverse-ratio
aggregation weights (Equation~\ref{eq:inverse_ratio_routing_weights}).
Routing follows the family-specific protocol in
\S\ref{subsec:baselines_protocol}.
All final heads use $\beta$-NLL. Prediction follows the joint sampling
procedure in \S\ref{subsec:aggregation_prediction}.
Table~\ref{tab:family_forecasting_results} reports means and seed
variability over the seven incoming datasets and seeds 40, 41 and 42.
Appendix~\ref{app:family_forecasting_details} gives prediction details and forecast examples.

\begin{table}[!htbp]
\centering
\footnotesize
\setlength{\tabcolsep}{3pt}
\renewcommand{\arraystretch}{1.12}
\caption{\footnotesize\textbf{Forecasting results and stored parameters.}
Mean $\pm$ sample SD over three seeds, each averaged over seven datasets.
Counts include the full encoder collection and forecasting head.
Bold marks the best mean or smallest count.}
\label{tab:family_forecasting_results}
\begin{tabular}{@{}lrrrrrr@{}}
\hline
Family & MSE $\downarrow$ & ECE $\downarrow$ & NLL $\downarrow$
& \shortstack{PICP90 (\%)\\$\rightarrow 90\%$}
& MPIW90 & Params (M) $\downarrow$ \\
\hline
SBNN
& $0.3896\pm0.0031$
& $\mathbf{0.0687}\pm0.0025$
& $\mathbf{0.814}\pm0.021$
& $\mathbf{90.47}\pm0.09$
& $1.771\pm0.018$
& \textbf{1.759} \\

Deterministic
& $0.4056\pm0.0008$
& $0.0816\pm0.0026$
& $1.353\pm0.422$
& $82.60\pm0.36$
& $1.476\pm0.007$
& 4.868 \\

MC Dropout
& $\mathbf{0.3894}\pm0.0041$
& $0.0767\pm0.0008$
& $0.817\pm0.004$
& $87.78\pm0.07$
& $1.647\pm0.007$
& 4.868 \\

Full-rank BBB
& $0.4426\pm0.0016$
& $0.0877\pm0.0018$
& $0.899\pm0.002$
& $92.29\pm0.06$
& $2.100\pm0.002$
& 9.731 \\

Deep ensemble (5)
& $0.3965\pm0.0028$
& $0.0734\pm0.0039$
& $0.881\pm0.008$
& $84.55\pm0.44$
& $1.517\pm0.007$
& 24.342 \\
\hline
\end{tabular}
\end{table}

\begin{figure}[tbp]
\centering
\begin{minipage}[t]{0.49\linewidth}
    \vspace{0pt}
    \centering
    \includegraphics[width=\linewidth]
    {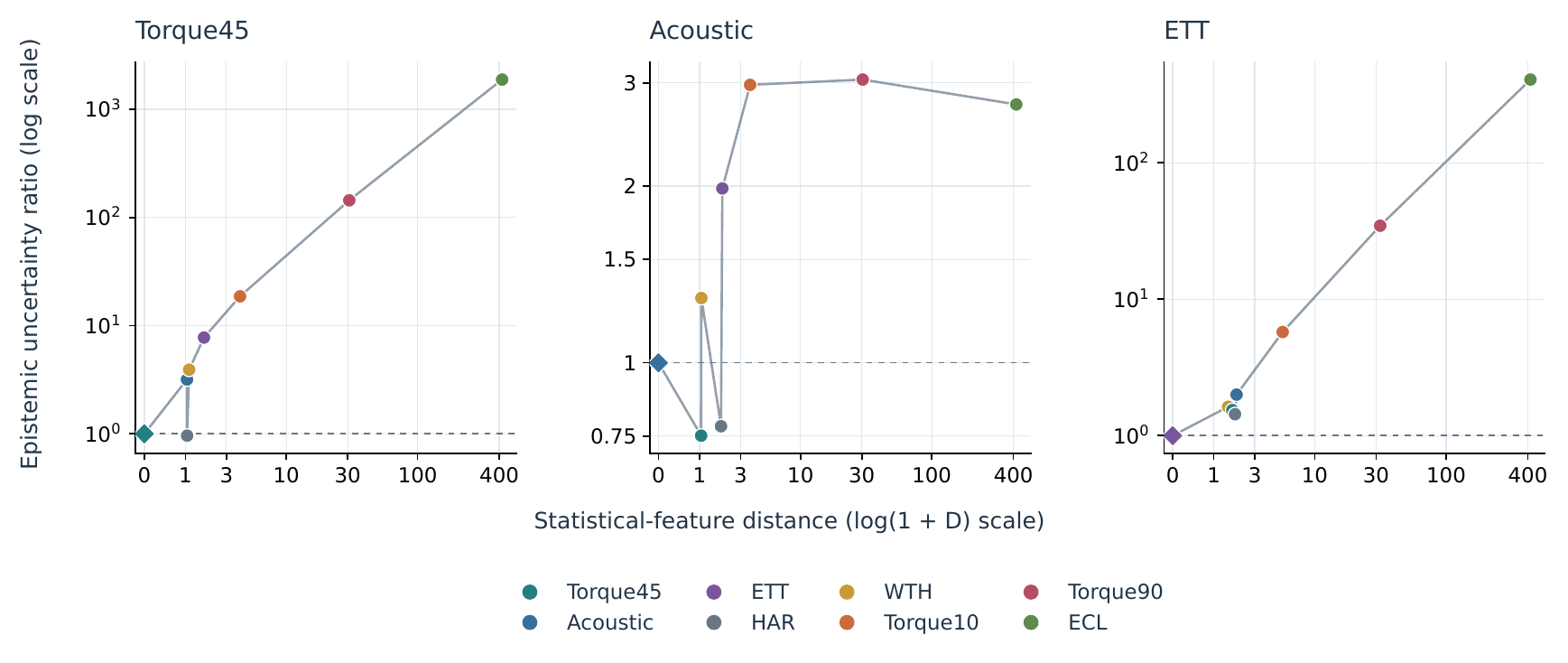}
    \caption{\textbf{Encoder uncertainty under shift.}
Colours denote datasets; diamonds mark sources.
Horizontal scale: $\log(1+D)$; vertical scale: logarithmic.
Dashed lines mark uncertainty ratio one.}
    \label{fig:encoder_shift_response}
\end{minipage}\hfill
\begin{minipage}[t]{0.49\linewidth}
    \vspace{0pt}
    \centering
    \includegraphics[width=\linewidth]
    {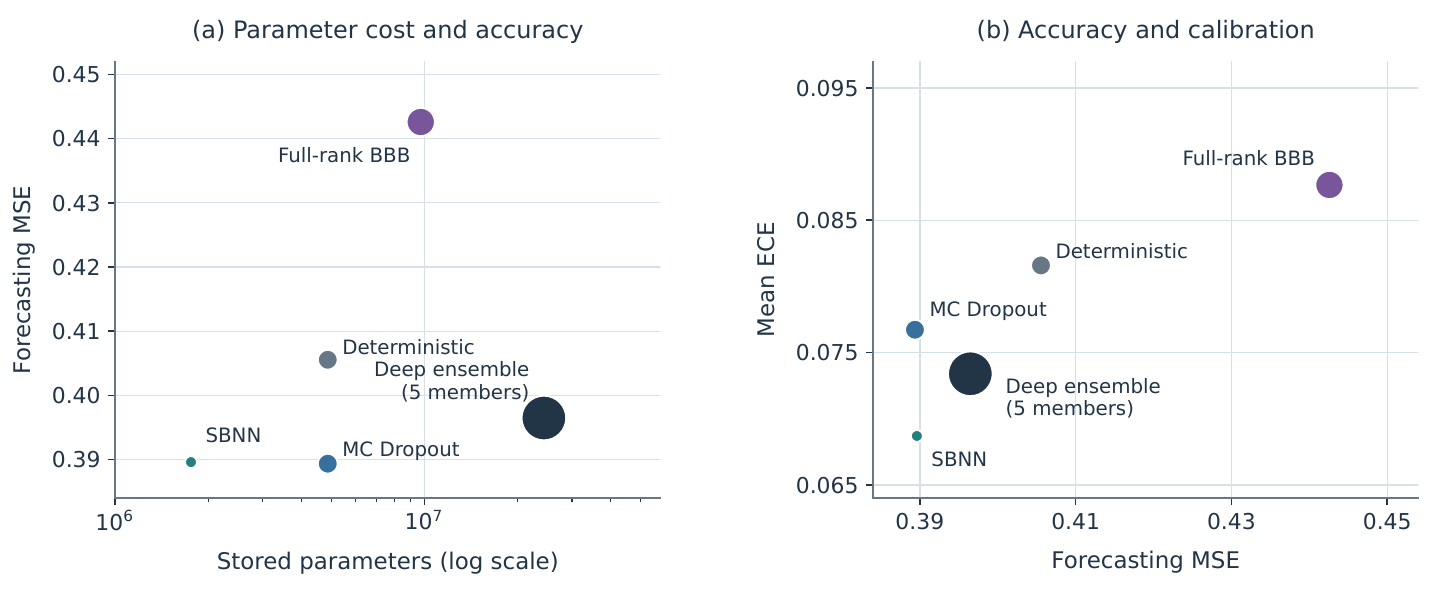}
    \caption{\textbf{Forecasting accuracy, calibration and cost.}
Seven-dataset means over three seeds: MSE versus stored encoder--head
parameters (left, log scale) and ECE (right).
Circle area is proportional to parameter count.}
    \label{fig:family_forecasting_scatters}
\end{minipage}
\end{figure}

\paragraph{Forecasting accuracy and storage.}
SBNN retains competitive forecasting accuracy after compressing
the encoders and the head.
With the forecasting head included, SBNN uses 63.9\% fewer
parameters than the deterministic and MC Dropout models,
81.9\% fewer than BBB and 92.8\% fewer than the ensemble.
At equal numerical precision, the same reductions apply to
parameter storage. Appendix~\ref{app:posterior_storage}
gives the encoder and head counts separately.
Figure~\ref{fig:family_forecasting_scatters} compares forecasting MSE, ECE and parameter count.
SBNN remains close to the best forecasting MSE while using
the fewest parameters.

\paragraph{Predictive calibration and coverage.}
The intervals and calibration metrics use the total predictive
variance in Equation~\ref{eq:total_predictive_variance}.
The learned $\sigma$ contributes to interval coverage in every
family, including the deterministic pipeline, whose weights remain fixed.
SBNN has the lowest mean ECE and NLL, and coverage closest to
the nominal 90\% level, although its NLL difference from MC Dropout
is small. Coverage must be considered together with interval width:
SBNN's improved coverage comes with wider intervals than the
under-covering MC Dropout and ensemble models, and narrower
intervals than BBB.

\begin{figure}[tbp]
\centering
\begin{minipage}[t]{0.54\linewidth}
    \vspace{0pt}
    \centering
    \includegraphics[width=\linewidth]
    {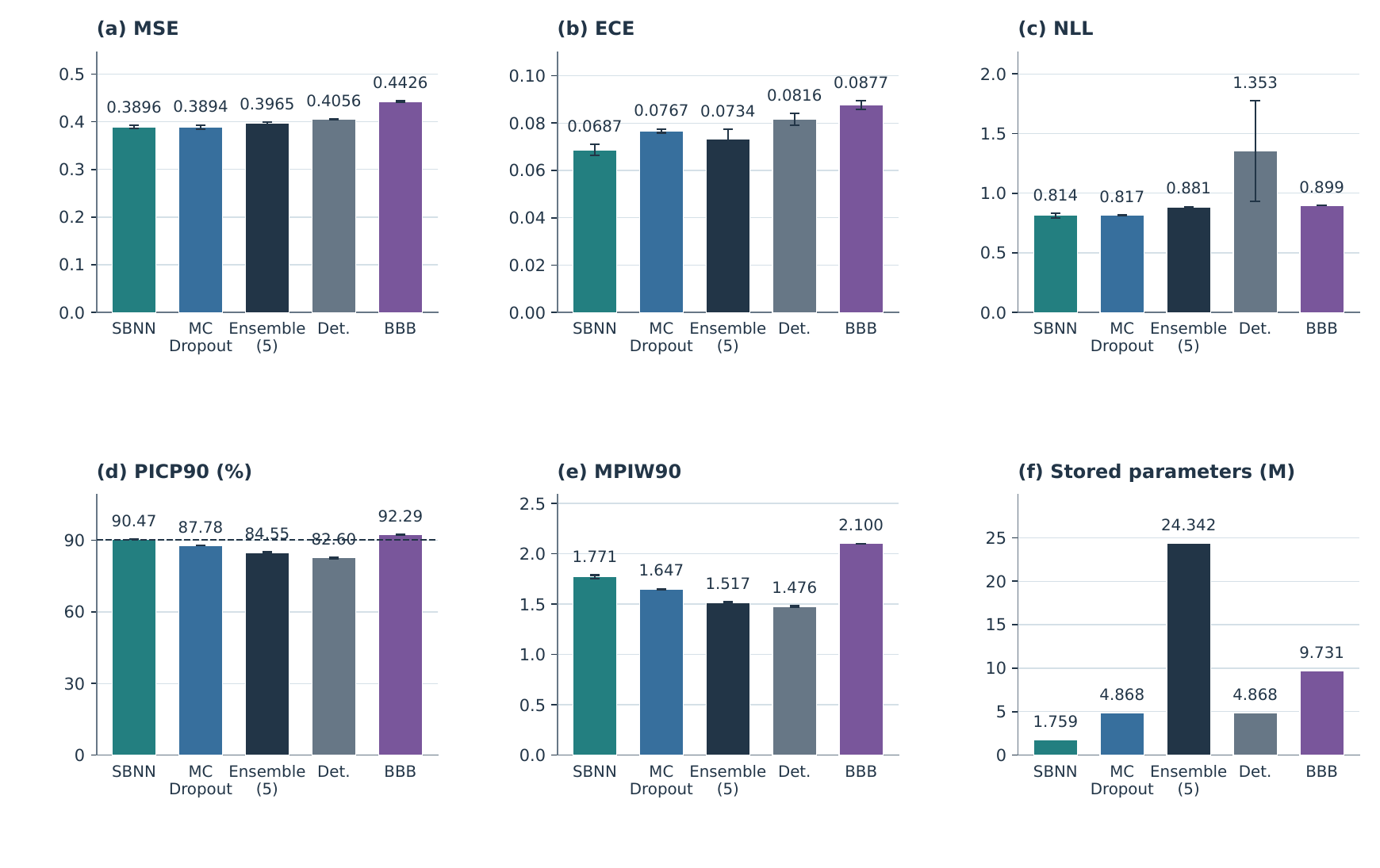}
\end{minipage}\hfill
\begin{minipage}[t]{0.44\linewidth}
    \vspace{0pt}
    \centering
    \includegraphics[width=\linewidth]
    {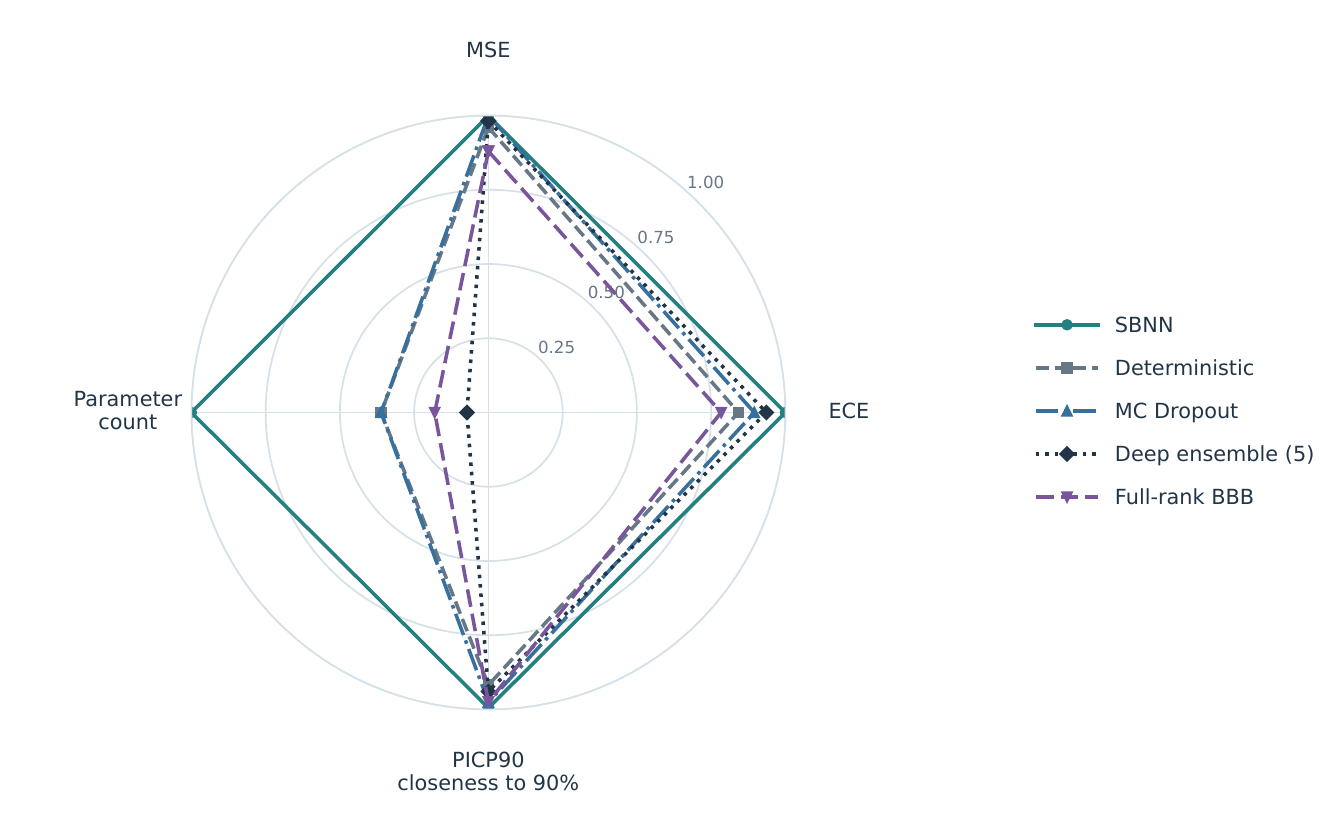}
\end{minipage}

\caption{\textbf{Forecasting quality and compactness.}
Top-4 uncertainty routing; seven datasets, three seeds.
Left: raw metrics and encoder--head counts; error bars show sample SD
of dataset-averaged seed results. Dashed line: 90\% coverage.
Right: higher is better. MSE, ECE and parameter scores use the lowest
family value divided by each family's value; coverage uses
$1-|\mathrm{PICP90}-0.90|/0.90$.}
\label{fig:family_forecasting_radar_bars}
\label{fig:family_forecasting_radar}
\end{figure}

Figure~\ref{fig:family_forecasting_radar_bars} summarizes forecasting
quality and parameter cost, including seed variability.

Appendix~\ref{app:family_forecasting_figures} compares observed
coverage with nominal levels of 70\% and 90\%.

Forecast examples and nominal 90\% prediction bands for all seven
incoming datasets are shown in
Figure~\ref{fig:family_forecasting_windows} in
Appendix~\ref{app:family_forecasting_details}.

\paragraph{Remark.}
The MSE differences between SBNN, MC Dropout and deep ensembles
are small. SBNN has slightly lower mean ECE and coverage closer
to the nominal 90\% level. These gains are modest.
The main advantage is that SBNN maintains competitive forecasting
accuracy and uncertainty quantification with substantially fewer
parameters and lower storage cost.

Appendix~\ref{app:initial_forecasting} reports the initial
MSE-trained comparison using epistemic uncertainty alone.

\subsection{Downstream comparison of uncertainty and distance routing}
\label{subsec:routing_comparison}

We now compare uncertainty-based and distance-based expert selection through downstream forecasting. Both policies select three SBNN encoders from the same eligible pool, using incoming inputs without forecast targets. The source-distance baseline selects the encoders with the lowest distances to their pretraining domains and assigns normalized inverse-distance weights. Uncertainty routing selects the encoders with the lowest source-normalized epistemic ratios and assigns normalized inverse-ratio weights. The comparison therefore evaluates each routing policy as a whole, including selection and weighting.

Both routing policies are training-free and label-free.
The distance baseline can cache its source statistical descriptors;
uncertainty routing stores a scalar source reference per encoder.

We compare top-3 distance with top-3 uncertainty routing because
adding a fourth distance-selected encoder reduced forecasting
accuracy and worsened calibration. Uncertainty routing with three
or four encoders gave similar results on most incoming datasets.
Using three experts in both policies therefore keeps the comparison
matched in the number of selected encoders.

Both policies use the matched head-training and prediction
protocol described in \S\ref{subsec:baselines_protocol}.
We evaluate seven incoming datasets with seeds $40$, $41$ and $42$,
giving 21 paired comparisons.

Table~\ref{tab:downstream_routing_comparison} in
Appendix~\ref{app:routing_results} reports the aggregate results. Means give equal weight to the seven datasets and
three seeds; the table also reports variability across seeds.
Uncertainty routing achieves a mean MSE of 0.3976, compared with
0.4062 for distance routing, a reduction of 2.1\%.
The MSE changes vary across datasets, with both improvements and
increases. The main advantage concerns predictive uncertainty.

\begin{figure}[!htbp]
\centering
\includegraphics[width=0.90\linewidth]
{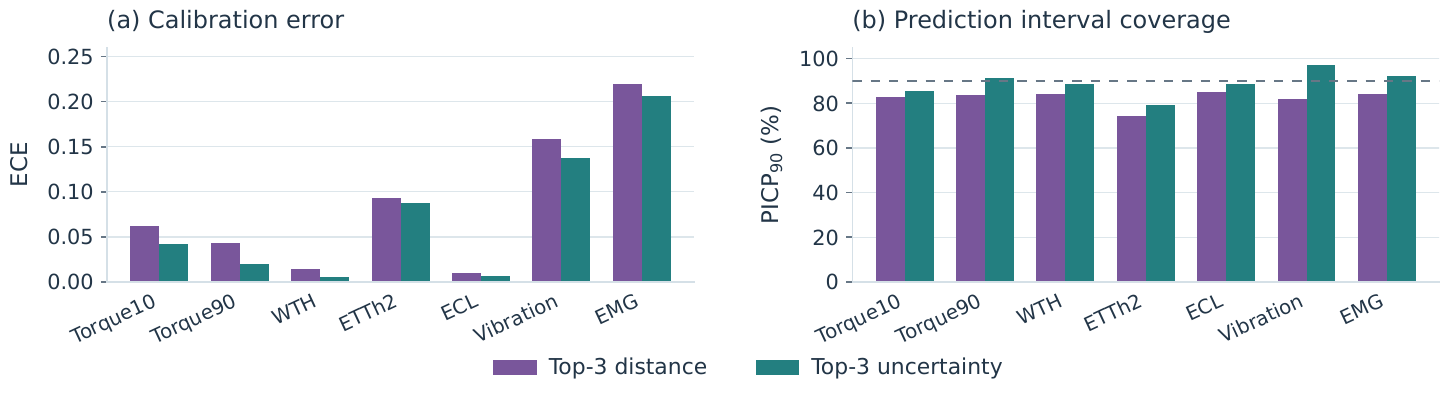}
\caption{\footnotesize\textbf{Calibration under distance and uncertainty routing.}
Bars average three seeds per dataset; the dashed line marks
nominal 90\% coverage.}
\label{fig:routing_distance_uncertainty_calibration}
\end{figure}

Mean ECE falls from 0.086 to 0.072 and mean NLL from 0.907 to 0.836.
Both metrics improve on all seven datasets after averaging over seeds.
Across individual runs, ECE improves in 19 of the 21 comparisons
and NLL in 20.
Figure~\ref{fig:routing_distance_uncertainty_calibration} shows
the per-dataset ECE and coverage results.

Mean PICP90 increases from 82.28\% to 89.03\%, a gain of
6.75 percentage points. Coverage is closer to the nominal 90\%
level in every paired dataset--seed comparison.
The figure also shows the remaining undercoverage and overcoverage.
Mean interval width decreases from 1.795 to 1.715:
the coverage improvement is accompanied by slightly narrower
intervals on average.

With the same number of selected experts, uncertainty routing retains competitive forecasting accuracy and improves predictive calibration and likelihood.
Per-dataset summaries, together with the MSE, NLL
and interval-width plots, are provided in
Appendix~\ref{app:routing_results}.

\subsection{Does uncertainty ranking predict downstream utility?}
\label{subsec:reconstruction_forecasting}

Uncertainty routing selects experts before the forecasting head is
trained. We compare the nine policies from \S\ref{subsec:baselines_protocol} on reconstruction and forecasting, using SBNN encoders throughout.

In \S\ref{subsec:source_reconstruction}, each encoder was evaluated
on its own source domain. Here, the frozen encoders and their decoders
reconstruct the seven incoming datasets. The MSE, ECE and coverage below describe reconstruction on the
seven incoming datasets.

\paragraph{Reconstruction favors the most confident experts.}
Table~\ref{tab:policy_reconstruction_results} reports the average
reconstruction metrics. Each policy combines individual encoder--decoder
errors as defined in Equation~\ref{eq:policy_reconstruction_metric}.

\begin{table}[!htbp]
\centering
\begin{minipage}[t]{0.41\linewidth}
\vspace{0pt}
\centering
\footnotesize
\setlength{\tabcolsep}{6pt}
\renewcommand{\arraystretch}{1.12}
\caption{\footnotesize\textbf{Average reconstruction metrics on seven incoming datasets.}}
\label{tab:policy_reconstruction_results}
\setlength{\tabcolsep}{2pt}
\renewcommand{\arraystretch}{1.04}
\begin{tabular}{@{}lrrr@{}}
\hline
Policy & MSE $\downarrow$ & ECE $\downarrow$
& \shortstack{PICP90 (\%)\\$\rightarrow 90\%$} \\
\hline
Top-3 uncertainty
& \textbf{0.0114} & \textbf{0.2777} & \textbf{63.10} \\
Top-4 uncertainty
& 0.1279 & 0.3123 & 56.04 \\
\hline
Source-diverse top-3
& 0.3647 & 0.3863 & 35.24 \\
Source-diverse top-4
& 0.3579 & 0.3641 & 39.83 \\
Random-4 of top-8
& 0.3926 & 0.3754 & 37.94 \\
Bottom-4 of top-8
& 0.3919 & 0.3850 & 41.09 \\
\hline
Random-4, full pool
& 0.5484 & 0.4065 & 28.91 \\
Bottom-3 uncertainty
& 0.4473 & 0.3485 & 39.28 \\
Bottom-4 uncertainty
& 0.5201 & 0.3763 & 33.80 \\
\hline
\end{tabular}
\end{minipage}\hfill
\begin{minipage}[t]{0.57\linewidth}
\vspace{0pt}
\centering
\footnotesize
\setlength{\tabcolsep}{4pt}
\renewcommand{\arraystretch}{1.12}
\caption{\footnotesize\textbf{Average forecasting metrics on seven incoming datasets.}}
\label{tab:policy_downstream_utility}
\setlength{\tabcolsep}{2pt}
\renewcommand{\arraystretch}{1.04}
\begin{tabular}{@{}lrrrrr@{}}
\hline
Policy & MSE $\downarrow$ & ECE $\downarrow$ & \shortstack{PICP90 (\%)\\$\rightarrow 90\%$} & NLL $\downarrow$ & MPIW90 \\
\hline
Top-3 uncertainty
& 0.3976 & 0.0721 & 89.03 & 0.8357 & 1.715\\
Top-4 uncertainty
& 0.3896 & 0.0687 & \textbf{90.47} & 0.8135 & 1.771 \\

\hline
Source-diverse top-3
& 0.3797 & 0.0682 & 91.02 & 0.7909 & 1.789 \\
Source-diverse top-4
& \textbf{0.3760} & 0.0682 & 91.02 & 0.7893 & 1.779 \\
Random-4 of top-8
& 0.3787 & \textbf{0.0676} & 90.61 & 0.8011 & 1.767 \\
Bottom-4 of top-8
& 0.3791 & 0.0702 & 90.98 & \textbf{0.7866} & 1.808 \\
\hline
Random-4, full pool
& 0.5085 & 0.0883 & 80.86 & 0.9002 & 2.069 \\
Bottom-3 uncertainty
& 0.4036 & 0.0848 & 87.52 & 0.8379 & 1.871 \\
Bottom-4 uncertainty
& 0.4038 & 0.0849 & 87.17 & 0.8458 & 1.861 \\
\hline
\end{tabular}
\end{minipage}
\end{table}

Top-3 and top-4 have the lowest mean reconstruction MSE and ECE,
and coverage closest to $90\%$. Top-4 achieves an MSE of $0.1279$,
compared with $0.3926$ for random-4 and $0.3919$ for bottom-4 from
the shortlist. Its ECE is also lower, and its coverage is closer
to nominal. The confidence ranking therefore works well for
reconstruction. Mean coverage nevertheless remains below $90\%$
on these incoming datasets.

\paragraph{Forecasting separates the shortlist from the controls.}
Table~\ref{tab:policy_downstream_utility} reports the corresponding
forecasting results. The low-uncertainty shortlist contains eight
encoders in this experiment. Top-3, top-4, random-4 of top-8 and
bottom-4 of top-8 all select within this shortlist.
Random-4 from the full pool and the bottom-3 and bottom-4
uncertainty policies provide the controls.

The shortlist policies have lower mean forecasting MSE, ECE and NLL
than all three controls. The clearest separation is in calibration
and coverage: their ECE ranges from $0.0676$ to $0.0721$, with
PICP90 between $89.03\%$ and $90.98\%$. The controls have ECE
between $0.0848$ and $0.0883$, and PICP90 between $80.86\%$
and $87.52\%$. Random-4 from the full pool also produces wider
intervals despite its lower coverage. The MSE gap is larger for
this random policy and smaller for the bottom-ranked controls.

\paragraph{The reconstruction ranking does not transfer.}
Within the shortlist, random-4 and bottom-4 have lower mean
forecasting MSE than top-3 and top-4, despite their higher
reconstruction errors.
Source-diverse top-4 gives the lowest forecasting MSE overall,
although its reconstruction MSE is higher than top-4's.

The differences within the shortlist are modest, and no policy
leads every forecasting metric. Random-4 has the lowest ECE,
bottom-4 has the lowest NLL, and top-4 has coverage closest to
$90\%$. Forecasting therefore shows a useful separation between
candidate groups, without preserving the reconstruction ordering
within the low-uncertainty group.

Figure~\ref{fig:policy_reconstruction_forecasting} in
Appendix~\ref{app:policy_results} compares the two tasks
using the same policy order.

Pretext-side encoder uncertainty helps exclude weak candidates.
Within the shortlist, reconstruction and forecasting favor different combinations, and their calibration rankings also differ.
Reconstruction averages individual encoder errors; forecasting
evaluates combined representations through a trained head.
We compare complete policies, including their aggregation weights. Detailed results are given in Appendix~\ref{app:policy_results}. Appendix~\ref{app:ratio_threshold_routing} also compares fixed-$K$ selection with a fixed uncertainty-ratio cutoff.
\section{Discussion and Conclusion}
\label{sec:discussion}

We presented a modular Bayesian framework for time series foundation models
that addresses the two main issues in expanding encoder collections:
storage cost and expert selection for incoming datasets.
Building on SBNN, we trained compact Bayesian encoders whose
epistemic uncertainty supports distribution-shift assessment
and label-free routing.
The resulting pipeline maintains competitive forecasting
accuracy and uncertainty calibration while requiring
substantially fewer stored parameters than the deterministic,
MC Dropout, full-rank Bayes by Backprop, and five-member
deep-ensemble pipelines. 
At the selected ranks, our Bayesian encoders require 63.8\% less
parameter storage than their deterministic counterparts.

The source-task results show that compactness can be achieved
alongside improved uncertainty calibration, with an
accuracy--calibration tradeoff that depends on the source
and selected rank and KL-scale configurations.
On unfamiliar inputs, encoder uncertainty increases with
measured source distance, supporting its use for shift assessment.
Uncertainty routing remains competitive with source-distance routing in forecasting MSE while improving mean ECE, NLL and agreement with nominal coverage.
These decisions use incoming inputs without target labels,
router training or encoder retraining.

The policy comparison also identifies a limit of this selection.
The most confident experts reconstruct incoming data better. Within the low-uncertainty shortlist, random, source-diverse or bottom-ranked combinations can outperform the top-ranked experts in forecasting.
Policies using the full pool or its least confident experts
generally give poorer forecasting results and uncertainty
calibration. Encoder uncertainty therefore helps narrow the
choice to a useful subset without determining the best combination within that subset for the
downstream task. In our experiments, better pretext
reconstruction does not guarantee better forecasting.

Several limitations remain.
Compression depends on the retained rank profiles, while rank
and KL scale require tuning to balance reconstruction and
calibration. The uncertainty response can also saturate under
large shifts, reducing its ability to distinguish very distant domains. We use a fixed projection dimension of 64; its effect on forecasting accuracy and calibration remains to be studied.
Final forecasting intervals combine epistemic uncertainty
with the head's learned aleatoric contribution.
Including this contribution improves coverage across model
families and narrows the larger calibration differences observed
in the initial epistemic-only comparison.
The final gains in forecasting calibration are therefore modest;
the strongest advantage is maintaining predictive quality at
substantially lower storage cost.
Our evaluation remains limited to the studied source
domains, incoming datasets and tasks.

The separation between pretext and downstream performance
connects this empirical finding to a broader theoretical question.
\citet{NEURIPS2021_02e656ad} establish conditions under which
reconstruction-based learning produces useful downstream
representations, including approximate conditional independence
and an informative reconstruction target.
\citet{pmlr-v313-wasserer26a} relate joint-embedding objectives
to the risk of an optimal downstream linear classifier.

Extending such results to compact Bayesian representations
would require accounting for the rank constraint and posterior
learning, as well as the relationship between the pretext and downstream tasks.

Future work could build on these analyses by designing pretext
objectives around the information required downstream.
The goal is to establish conditions under which learning a
pretext task bounds downstream risk.
Masked reconstruction and forecasting provide one setting
for this investigation; other pretext and downstream tasks
raise the same question.
Such guarantees would give a theoretical basis for deciding
when a compact Bayesian representation can be reused
successfully, beyond the empirical shortlisting criterion
established here.
\newpage
\bibliography{main}
\bibliographystyle{tmlr}
\newpage
\appendix

\section{Related Work}
\label{sec:related_work}

\subsection{Modular models and expert selection}

Modular foundation models reuse pretrained components across
datasets and tasks~\citep{Pfeiffer2023Modular,Pham2026Modular}.
This allows new specialists to be added while retaining the
components already trained. Reuse then requires deciding how to combine knowledge acquired on different sources.

\citet{syed2026position} examine why choosing models by aggregate
benchmark scores can be unreliable. They evaluate seven time-series
foundation models on 72 datasets spanning six application domains
and find that model rankings vary across domains and datasets.
They call for domain-stratified evaluation and methods for assessing
which source domains help or harm transfer to a target domain.
This provides empirical motivation for retaining specialised encoders
and selecting them according to incoming data. Their study evaluates
zero-shot point forecasting; here, we study how an encoder collection
can remain compact while providing an uncertainty signal for selection.

The reuse of pre-trained modules has been studied in language models.
AdapterFusion~\citep{Pfeiffer2021AdapterFusion} decouples the training
of task-specific adapters from a subsequent composition step, in
which an attention mechanism learns to combine their representations.

During composition, both the adapters and their shared
language-model backbone remain frozen. Our encoder collection
follows this separation between source training and downstream
reuse, with whole encoders as the reusable components.
\citet{Ostapenko2024Modular} build libraries of low-rank
adaptation (LoRA) adapters and
propose Arrow, a mechanism that selects relevant adapters for new
inputs without requiring additional training. Thus, routing within
a library of modules does not always require supervision on the
target task. Modularity can also be applied to heterogeneous
architectures; Manticore combines blocks from pre-trained models
using projections and learned mixing weights~\citep{Roberts2024Manticore}.
These methods differ in the components they reuse and the training needed to combine them.

Guard learns to combine forecasts from pretrained time-series models for distillation into a compact student~\citep{Dey2026Guard}. Teacher forecasting errors provide supervision for its contextual router, while an uncertainty-dependent temperature controls distillation strength. We retain the encoders as reusable experts and compute their selection scores from unlabeled incoming windows.\par Selection also depends on what is known about the target task.
Log Expected Empirical Prediction (LEEP) uses labeled target examples to score the transferability
of a pretrained classifier~\citep{Nguyen2020}.
Our router uses unlabeled inputs.
For time series, \citet{Pham2026Modular} provide the closest
comparison to our encoder collection. Their self-supervised
encoders remain frozen during downstream adaptation. Selection
first uses statistical and spectral descriptors to measure
source--target dataset distance, then ranks candidates by how well
their representations preserve input similarities. Selected
representations are aligned and combined by a Transformer
aggregation module. We share this separation between pretraining,
selection, and task adaptation. Our selection score comes from
the Bayesian encoders' own uncertainty on incoming data.
Our downstream comparison between uncertainty based routing and distance based routing uses a nearest-source distance baseline
inspired by this first criterion. 
A different approach is to train expert selection together with
the experts.
The learned gate in \citet{jacobs1991moe} assigns inputs to experts
during training. Time-MoE and Moirai-MoE use sparse expert activation
within time-series models~\citep{shi2024timemoe,Liu2025MoiraiMoE}.
Our unit of selection is a whole pretrained encoder. We select a
fixed set for each incoming dataset, and new encoders can enter
the collection without retraining existing experts or a gate.
\subsection{Storage cost of Bayesian encoders}

The storage cost of an encoder collection grows as specialists
are added, even when the models are deterministic. The additional
storage required by conventional Bayesian neural networks has
long been an obstacle to scalable uncertainty
quantification~\citep{Dusenberry2020}. Our goal is to reduce the
storage cost of the collection while providing epistemic uncertainty for
distribution-shift assessment and expert selection.

Under full-rank mean-field Gaussian inference, each Bayesian weight
requires a mean and a scale. Bayes by Backprop learns
both~\citep{Blundell2015}. This doubles the stored parameters for
those weights relative to a deterministic parameterization.
Deep ensembles estimate uncertainty from several independently
trained networks~\citep{Lakshminarayanan2017}. Our five-member
baseline therefore stores five complete copies of the network.
MC Dropout uses stochastic forward passes through one network
and retains its deterministic weight matrices~\citep{Gal2016}.

Low-rank methods have been developed to address the scalability
of Bayesian deep learning~\citep{tomczak2020efficient,Dusenberry2020,wang2024blob}.
Stochastic weight averaging-Gaussian (SWAG) and structured Gaussian variational inference use compact
covariance representations while retaining full-dimensional weight
means~\citep{Maddox2019,tomczak2020efficient}.
Rank-1 BNNs learn a shared dense weight matrix and distributions
over multiplicative vectors~\citep{Dusenberry2020}.
The rank-one modulation reduces Bayesian overhead while retaining
the shared dense weight matrix. Bayesian Low-Rank Adaptation by Backpropagation (BLoB) learns a Bayesian
low-rank adapter to a frozen backbone~\citep{wang2024blob}.
These approaches reduce posterior or adaptation overhead while
retaining full-dimensional weight means, a shared dense matrix,
or a frozen backbone. SBNNs parameterize the layer weights
themselves with Bayesian low-rank factors~\citep{Toure2026}. We use this parameterization to build
Bayesian encoders that cost less to store than their deterministic
counterparts and provide epistemic uncertainty for expert selection.
Appendix~\ref{app:posterior_storage}
compares the stored components of these methods.

Bayesian-MoE applies a post-hoc Laplace approximation to the second linear layer of each expert in an MoE language model~\citep{Dialameh2025BayesianMoE}. Its low-rank Kronecker-factored approximation concerns posterior curvature around retained dense weights. The learned gate is kept, and the posterior supplies predictive uncertainty estimates within expert pathways.\par The original SBNN study reports comparisons with both Rank-1 BNNs
and SWAG~\citep{Toure2026}. Those experiments complement the
modular forecasting evaluation presented here.

\subsection{Uncertainty for routing and distribution-shift detection}

Our routing signal comes from the pretrained encoders before downstream training.

Mixture-of-Gaussians with Uncertainty-based Gating (MoGU) predicts a Gaussian mean and variance for each expert and
computes mixture weights in inverse proportion to those
variances~\citep{aviv2025mogu}. It requires no separate gating
network; the variances are learned with the forecasting objective.
\citet{Zhou2026ExpertsUncertainty} combine label-distribution learning
with multiple forecasting experts and an MLP gate. Their continuous
variant learns Gaussian-mixture forecasts by matching predicted and
enhanced target distributions with maximum mean discrepancy.
Evaluated on aggregated M5 sales data, this approach learns the
gate and predictive distributions for the forecasting task.

For sparse language models, \citet{Nahshan2026CrossEntropyRouting}
supervise routing with realized next-token cross-entropy.
One formulation learns expert-indexed error scores that modify
affinities before top-$K$ selection; another directly supervises
affinity concentration. The observed loss provides supervision during
training, while inference uses the learned routing mechanism without
requiring the true next token. Our router requires no such
target-task supervision or router fitting.

Uncertainty-Aware Routing (UAR) uses router entropy and free energy
derived from router logits to adapt expert activation and
load-balancing regularization~\citep{chen-etal-2026-uncertainty-aware}.
These signals describe the learned gate. Our score measures variation in each encoder's representations
under posterior weight draws. 

Variational Routing learns probabilistic expert assignments within pretrained MoE language models~\citep{Li2026VariationalRouting}. Its inference networks model routing-logit distributions or learn input-dependent temperatures for stochastic selection. \citet{CajasOrdonez2025QuantizedMoE} use MC Dropout in a learned router and increase the weights of higher-precision quantized experts when uncertainty rises. Both methods train routing components for their tasks. Our routing statistic is computed directly from posterior representation draws of frozen encoders and their stored source references.\par Grassmannian MoE represents experts through learned subspaces and
computes routing weights from subspace affinities and Bingham
concentration parameters~\citep{Shihab2026GrassmannianMoE}.
An amortization network allows token-dependent assignment distributions.
These low-dimensional subspaces parameterize the router, with the
expert weight matrices retained separately. GrMoE models uncertainty
in expert assignment. Our score measures representation variation
under the encoder weight posterior.

VI-MoLE learns the counterfactual residual risk of nested LoRA-expert
prefixes, calibrates risk estimates, and allocates a shared compute
budget using predicted risk reduction per unit
cost~\citep{Saliencro2026VIMoLE}. It also supports abstention.
A committee-only target can omit labels; task-risk supervision
and answer-risk calibration use labeled data. Its acquisition policy
is learned and calibrated, while our incoming-task routing scores
come directly from the frozen encoders and their stored source
references.

In image restoration, \citet{Zhang_2024} use MC Dropout within
a learned router to estimate and normalize its routing logits.
We select and weight pretrained encoders using their epistemic uncertainty before fitting the forecasting head.

Using an encoder's uncertainty to assess unfamiliar data also
connects routing to distribution-shift detection.
\citet{pmlr-v244-arnez24a} compare entropy statistics of latent
representations sampled through dropout or DropBlock with a
reference fitted on source data. 
We obtain uncertainty through the encoder's variational weight
posterior and use the source-normalized score to select and weight
experts as well as assess shift. The response to shift depends
on the model: \citet{dangelo2022outofdistributiondetectionbayesianneural}
show that even exact Bayesian inference can give poor OOD
detection for common neural-network priors. We therefore evaluate
the score against measured distribution shift and assess its
downstream usefulness separately.

\subsection{Downstream utility and calibration}
Ensemble selection makes the dependence on the target task explicit by
choosing combinations according to held-out
performance~\citep{caruana2004ensemble}. Individual accuracy also
does not determine the value of a combination. The analysis of
\citet{wood2023unified} relates ensemble error to bias, variance,
and diversity, while \citet{gontijolopes2022norepresentation} show
that models with complementary features can improve downstream
performance even when one is less accurate on its own.
These results motivate examining the selected group alongside
its individual encoders. In our setting, we compare reconstruction
and forecasting under the same policies to establish when a
selection that works for the pretext task also works downstream.

VI-MoLE separates current predictive uncertainty from the risk
reduction expected from acquiring another
expert~\citep{Saliencro2026VIMoLE}. In our experiments, uncertainty identifies a useful shortlist, and forecasting rankings vary within it. VI-MoLE analyzes task-specific expert acquisition. Our future work
asks when pretext learning gives guarantees on downstream performance.

Forecast error and predictive calibration answer different
questions~\citep{Kuleshov2018Calibration}.
\citet{Adler2026Calibration} assess calibration in time-series
foundation models, including its dependence on prediction heads
and forecast length. We report forecasting MSE for point accuracy and calibration error, NLL, and interval coverage for the predictive distribution.

\section{Dataset and evaluation protocol}
\label{app:dataset_protocol}

Incoming datasets are new to the encoder collection; they were not used during source pretraining. Dataset descriptions and origins are given in Section~\ref{app:dataset_descriptions}; routing settings are given in Appendix~\ref{app:routing_protocol_details}.\par \subsection{Dataset descriptions} \label{app:dataset_descriptions}

We use four source datasets to pretrain the encoders and seven
incoming datasets to evaluate their reuse.
Table~\ref{tab:dataset_descriptions} identifies the signals
and the role of each dataset.
Acoustic, the torque datasets and vibration contain recordings
collected by an industrial partner.

The ETT source encoder is trained on oil-temperature windows
from ETTh1. Incoming ETTh2 contains measurements from a different
transformer. Both belong to the hourly Electricity Transformer
Dataset~\citep{haoyietal-informer-2021}.

\begin{table}[H]
\centering
\small
\setlength{\tabcolsep}{5pt}
\renewcommand{\arraystretch}{1.15}
\caption{\textbf{Source and incoming datasets.}
Source datasets support encoder pretraining.
Incoming datasets support routing, reconstruction evaluation
and downstream forecasting.}
\label{tab:dataset_descriptions}
\begin{tabular}{@{}p{0.17\linewidth}p{0.45\linewidth}p{0.30\linewidth}@{}}
\hline
Dataset & Measured signal & Origin \\
\hline
\multicolumn{3}{l}{\textit{Source datasets}} \\
Acoustic
& Generator acoustic signal
& Industrial partner \\

HAR
& Human-activity signals
& Public~\citep{UCIHAR2013} \\

Torque45
& Shaft torque signal at 45\% of maximum unit power
& Industrial partner \\

ETT (ETTh1)
& Transformer oil temperature
& Public~\citep{haoyietal-informer-2021} \\
\hline
\multicolumn{3}{l}{\textit{Incoming datasets}} \\
Torque10
& Shaft torque signal at 10\% of maximum unit power
& Industrial partner \\

Torque90
& Shaft torque signal at 90\% of maximum unit power
& Industrial partner \\

WTH
& Weather measurements: wet-bulb temperature
& Public~\citep{haoyietal-informer-2021} \\

ETTh2
& Transformer oil temperature
& Public~\citep{haoyietal-informer-2021} \\

ECL
& Electricity consumption
& Public~\citep{Trindade2015Electricity} \\

Vibration
& Generator vibration signal
& Industrial partner \\

EMG
& Electromyographic signals
& {\raggedright Public (PhysioNet)~\citep{Goldberger2000PhysioNet}\par} \\
\hline
\end{tabular}
\end{table}
Each experiment uses a single signal channel.
For WTH, ECL and ETTh2, we retain the wet-bulb temperature,
MT\_320 electricity-consumption and oil-temperature channels,
respectively.
The stored EMG windows are concatenated in their saved order
to construct the series used for forecasting.

For all seven incoming datasets, the forecasting head predicts
the next 96 time steps from an input window of 96 past time steps.
For encoder reuse, each incoming window is linearly interpolated
on a normalized time grid to the encoder's native input length.

\section{Source pretraining objectives and settings}
\label{app:source_pretext_protocol}

\subsection{Encoder construction and training}
\label{subsec:bayesian_encoder_learning}
\paragraph{Pretext objective.}
The general objective combines reconstruction and contrastive learning.
Masked reconstruction trains the encoder and decoder to recover a clean
signal from an incomplete input, following the denoising autoencoder
principle~\citep{Vincent2008Denoising}.
For an input window $x$, a binary mask $m$ retains some values and
hides others. The encoder receives $\tilde{x}=m\odot x$, and the
decoder $g_{\mathrm{dec}}$ returns
$\hat{x}=g_{\mathrm{dec}}(f(\tilde{x}))$. We score reconstruction
against the clean signal over the whole window:
\begin{equation}
    \mathcal{L}_{\mathrm{rec}}
    = \frac{1}{d_x}
      \sum_{j=1}^{d_x}(x_j-\hat{x}_j)^2,
    \label{eq:source_reconstruction_loss}
\end{equation}
where $d_x$ is the number of scalar values in the window.
The decoder and its parameters are omitted from the weight
notation below.

Contrastive learning encourages similar representations for two
views of the same window and separates them from representations
of other windows in the batch. Views are constructed through input
augmentations, without class labels. We use cosine similarity between
normalized representations and a temperature parameter, following
\citet{Chen2020SimCLR}.
With $\mathcal{L}_{\mathrm{con}}$ denoting the contrastive loss,
the combined loss is
\begin{equation}
    \mathcal{L}_{\mathrm{SSL}}
    = (1-\alpha)\mathcal{L}_{\mathrm{rec}}
      +\alpha\tilde{\lambda}(t)\mathcal{L}_{\mathrm{con}},
    \label{eq:source_pretext_loss}\end{equation}
where $\alpha$ controls the mixture and $\tilde{\lambda}(t)$
balances the magnitudes of the two terms at training step $t$.
This scale is detached from gradients.
The contrastive variants, masking settings and loss coefficients
for each source are specified below.

\paragraph{Bayesian low-rank weights.}
Let $\ell$ index the layers.
The weight matrix $W_\ell$ has $a_\ell$ rows and $b_\ell$ columns,
and $r_\ell$ is our chosen rank.
SBNN writes
\begin{equation}
    W_\ell=A_\ell B_\ell^\top,
    \qquad
    A_\ell\in\mathbb{R}^{a_\ell\times r_\ell},
    \quad
    B_\ell\in\mathbb{R}^{b_\ell\times r_\ell}.
    \label{eq:encoder_lowrank_weights}
\end{equation}
Both factors are stochastic. We learn a Gaussian mean and scale
for each factor entry. For the factors,
$q(\Theta)=\prod_\ell q(A_\ell)q(B_\ell)$, with
$\Theta=\{A_\ell,B_\ell\}_\ell$.
Every sampled matrix has rank at most $r_\ell$.
Its entries share factors, so independent factor entries induce
dependencies between weights. Convolutional kernels are reshaped
into matrices for the same factorization.

A deterministic matrix requires
$a_\ell b_\ell$ parameters, and full-rank mean-field Bayesian
weights require $2a_\ell b_\ell$. SBNN stores
$2r_\ell(a_\ell+b_\ell)$ factor means and scales.
It is smaller than the deterministic matrix when
$2r_\ell(a_\ell+b_\ell)<a_\ell b_\ell$, and smaller than the
full-rank Bayesian parameterization when
$r_\ell(a_\ell+b_\ell)<a_\ell b_\ell$.
Biases and other parameters are counted separately.

\paragraph{Training and configuration choices.}
Following \citet{Toure2026}, we initialize the factor means from a truncated singular value
decomposition of the corresponding deterministic model's weights,
then learn the factor means and scales on the source task.
The Bayesian training objective is
\begin{equation}
    \mathcal{J}_{\mathrm{source}}(q)
    = \mathbb{E}_{q}\bigl[\mathcal{L}_{\mathrm{SSL}}\bigr]
      +\lambda_{\mathrm{KL}}(t)\mathcal{K}(q,p),
    \label{eq:encoder_training_objective}
\end{equation}
where $\mathcal{K}(q,p)$ sums the KL terms between the factor
posteriors $q(A_\ell)$ and $q(B_\ell)$ and their respective
priors $p(A_\ell)$ and $p(B_\ell)$. Other Bayesian parameters,
where present, contribute their own KL terms. The coefficient
$\lambda_{\mathrm{KL}}(t)$ sets their strength at training step
$t$.  We estimate the expected loss with weight samples and
differentiate through Gaussian reparameterization~\citep{Kingma2014AutoEncoding}.

The rank profile determines storage cost. Lower ranks use fewer
parameters but restrict the weight maps. We treat rank as a hyperparameter. Through rank ablations,
we select the smallest tested rank that remains competitive
with the deterministic counterpart on the validation metric
of interest, within the storage budget. The reported compression
therefore depends on the ranks selected for each encoder. The KL scale changes
posterior regularization without changing that count. A storage
budget can therefore constrain the admissible ranks, while rank
and KL settings are chosen for the chosen validation metric.
In our source comparisons, \emph{SBNN (reconstruction)} prioritizes
low validation MSE. \emph{SBNN (calibration)} prioritizes low ECE
and coverage close to its nominal level. \emph{SBNN (balanced)}
considers reconstruction and calibration together. These labels identify selection priorities within the SBNN family.
Rank and KL settings vary by encoder.
Appendix~\ref{app:rank_kl_sweep} shows how the tested rank
combinations and KL scales affect reconstruction performance
and calibration on ETT.

The pretext loss and Bayesian objective are defined in
Equations~\ref{eq:source_pretext_loss} and
\ref{eq:encoder_training_objective}.
The contrastive and KL coefficients vary across retained configurations.
For SBNN, KL regularization acts on the factor posteriors
\citep{Toure2026}; for full-rank BBB, it acts on the weight
posteriors~\citep{Blundell2015}.
Deterministic and MC Dropout training omit this variational
KL term, as does each independently trained ensemble member.
\paragraph{Encoder-specific coefficients.}
Table~\ref{tab:source_pretext_settings} gives the retained
SBNN training coefficients and the full-rank BBB coefficients
used for each source family.
SBNN configurations with identical coefficients are grouped.

\begin{table}[H]
\centering
\small
\setlength{\tabcolsep}{6pt}
\renewcommand{\arraystretch}{1.12}
\caption{\textbf{Source-pretraining coefficients.}
$\alpha$ controls the reconstruction--contrastive mixture.
KL columns give the final regularization coefficients.
Full-rank BBB coefficients are specified per source family.}
\label{tab:source_pretext_settings}
\begin{tabular}{@{}lccc@{}}
\hline
SBNN configuration & $\alpha$
& \shortstack{KL coefficient\\SBNN}
& \shortstack{KL coefficient\\Full-rank BBB} \\
\hline
Acoustic balanced
& 0.1 & $10^{-4}$ & $5\times10^{-6}$ \\
Acoustic lean
& 0.1 & $7.5\times10^{-5}$ & $5\times10^{-6}$ \\
Acoustic counterpart
& 0.1 & $1.5\times10^{-4}$ & $5\times10^{-6}$ \\
Acoustic strong variants (3 encoders)
& 0.1 & $5\times10^{-6}$ & $5\times10^{-6}$ \\
Acoustic CNN
& 0 & $3\times10^{-4}$ & $5\times10^{-6}$ \\
\hline
HAR variants (3 encoders)
& 0.5 & $0.03$ & $0.03$ \\
\hline
Torque45 variants (3 encoders)
& 0 & $5\times10^{-5}$ & $10^{-3}$ \\
\hline
ETT base (KL $5\times10^{-7}$)
& 0.5 & $5\times10^{-7}$ & $10^{-6}$ \\
ETT base (KL $10^{-6}$)
& 0.5 & $10^{-6}$ & $10^{-6}$ \\
ETT wider
& 0.5 & $2\times10^{-6}$ & $10^{-6}$ \\
ETT compact
& 0.5 & $10^{-6}$ & $10^{-6}$ \\
\hline
\end{tabular}
\end{table}

\paragraph{Masking settings.}
Table~\ref{tab:source_masking_settings} gives the input-masking
settings used during source pretraining.

\begin{table}[H]
\centering
\small
\renewcommand{\arraystretch}{1.12}
\caption{\textbf{Source-pretraining masking settings.}
Point and span masking are separate operations.
A zero ratio means that no reconstruction mask is applied.}
\label{tab:source_masking_settings}
\begin{tabular}{@{}ll@{}}
\hline
Encoder configuration & Configured masking \\
\hline
Acoustic Transformer & Point fraction 0.05; span fraction 0.25 \\
HAR                  & Mask ratio 0.50 \\
ETT                  & Mask ratio 0.25 \\
Torque45             & Mask ratio 0 \\
Acoustic CNN         & Mask ratio 0 \\
\hline
\end{tabular}
\end{table} 

\paragraph{Contrastive settings.}
The contrastive losses use a temperature of $0.2$.
Acoustic Transformer encoders use symmetric cross-view
InfoNCE; HAR and ETT encoders use a polynomial variant
of NT-Xent.
The loss-balancing factor is detached from gradients.
For Acoustic, it is updated during training using an
exponential moving average.
For HAR and ETT, it is computed from the first batch
and then held fixed.
\section{Rank and KL ablation on ETT}
\label{app:rank_kl_sweep}

\begin{figure}[H]
    \centering
    \includegraphics[width=\linewidth]{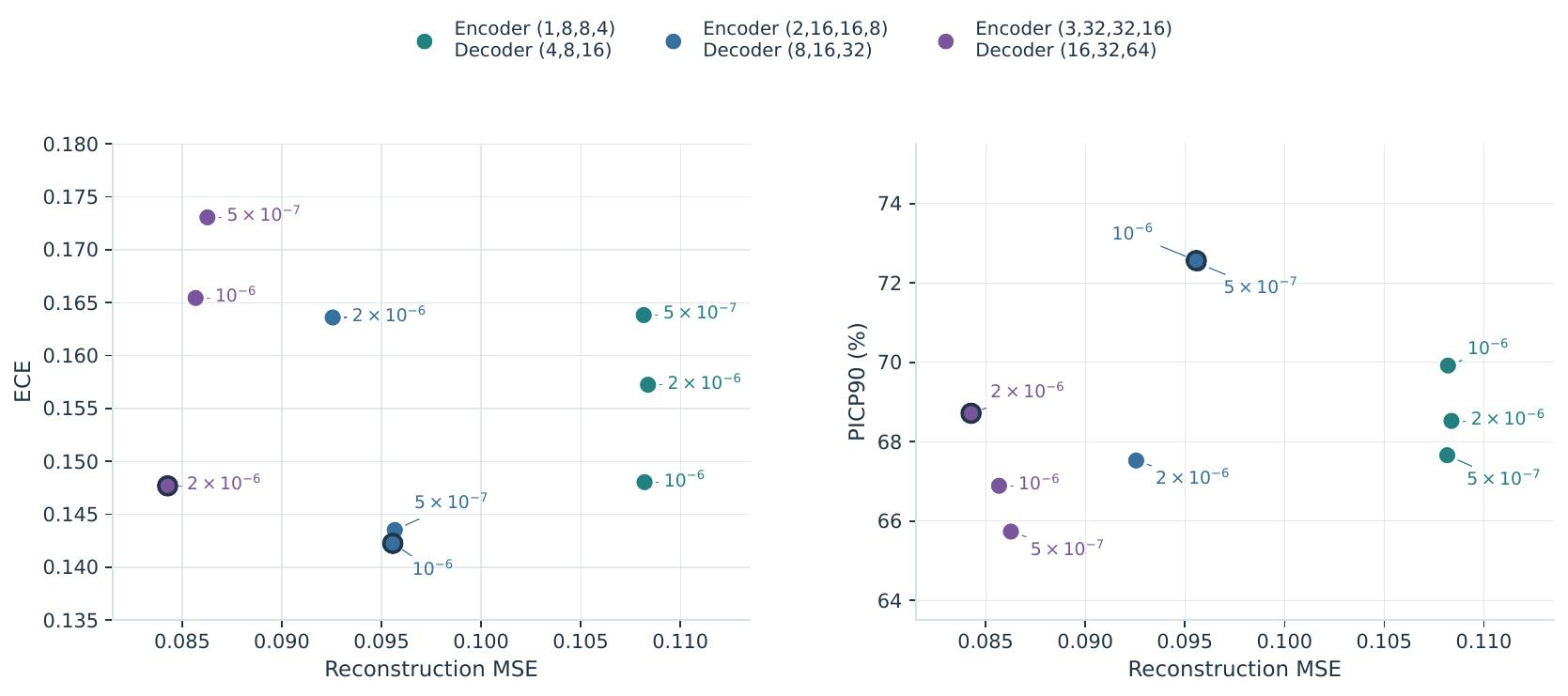}
    \caption{\textbf{Rank--KL tradeoffs in ETT source reconstruction.}
    Each circle represents one tested configuration. Colours identify
    the encoder and decoder ranks; point labels give the KL scale.
    Encoder ranks follow the order convolution 1, convolution 2,
    intermediate projection, and latent layer. Decoder ranks follow
    the order fully connected layer 1, fully connected layer 2,
    and output layer. MSE is evaluated over the whole window;
    ECE and PICP90 use the masked-point calibration evaluation.
    Dark outlines mark non-dominated configurations for MSE and ECE
    (left), or MSE and absolute coverage error from 90\% (right).}
    \label{fig:ett_rank_pareto}
\end{figure}

We compare three joint encoder-decoder rank profiles at three
KL scales, giving nine configurations with seed 40.
The posterior scale settings are fixed across candidates.
Figure~\ref{fig:ett_rank_pareto} shows their reconstruction MSE,
ECE, and PICP90. Increasing rank reduces reconstruction MSE
in this sweep, while calibration and coverage depend on both
rank and KL scale. The configuration with the lowest MSE
does not have the lowest ECE or the coverage closest to 90\%.

\section{Evaluation metrics}
\label{app:evaluation_metrics}

\paragraph{Reconstruction.}
We evaluate each selected encoder with its decoder on the incoming
dataset. Let $R_m$ denote its reconstruction MSE, averaged over
input windows and their coordinates.
For the selected set of encoders $\mathcal{S}$, the policy score is
\begin{equation}
    R_{\mathcal{S}}
    = \sum_{m\in\mathcal{S}}\alpha_m R_m,
    \label{eq:policy_reconstruction_metric}
\end{equation}
where $\alpha_m$ is the weight assigned to encoder $m$ by the
evaluated policy.
We compute each encoder's reconstruction MSE separately, then take
their weighted average.
Policies are compared within each incoming dataset, since raw
errors from signals with different scales are not directly comparable.

\paragraph{Forecasting accuracy.}
Let $N_{\mathrm{test}}$ be the number of test windows and $H$ the
number of forecast coordinates per window. Using the predictive mean
$\widehat{\mu}$ from \S\ref{subsec:aggregation_prediction}, we compute
\begin{equation}
    \mathrm{MSE}
    = \frac{1}{N_{\mathrm{test}}H}
      \sum_{i=1}^{N_{\mathrm{test}}}\sum_{h=1}^{H}
      \left(y_{i,h}-\widehat{\mu}_{i,h}\right)^2.
    \label{eq:forecast_mse}
\end{equation}
Forecasting metrics use the standardized target scale.

\paragraph{Predictive negative log-likelihood.}
We evaluate a Gaussian approximation with predictive mean
$\widehat{\mu}$ and total variance $\widehat{v}_{\mathrm{pred}}$.
For numerical stability, let
$v_{i,h}=\max\{\widehat{v}_{\mathrm{pred},i,h},10^{-12}\}$.
The reported negative log-likelihood is
\begin{equation}
    \mathrm{NLL}
    = \frac{1}{2N_{\mathrm{test}}H}
      \sum_{i=1}^{N_{\mathrm{test}}}\sum_{h=1}^{H}
      \left[
          \log(2\pi v_{i,h})
          +\frac{(y_{i,h}-\widehat{\mu}_{i,h})^2}{v_{i,h}}
      \right].
    \label{eq:predictive_nll}
\end{equation}
Lower NLL indicates a better predictive density at the observed
targets. This evaluation metric uses the total predictive variance
and is distinct from the $\beta$-NLL training objective.

\paragraph{Calibration and interval coverage.}
For downstream forecasting, we construct pointwise Gaussian
prediction intervals using
$\widehat{\sigma}=\sqrt{\widehat{v}_{\mathrm{pred}}}$.
For nominal coverage $q$, the interval is
\begin{equation}
    I_{i,h}(q)
    =\left[
        \widehat{\mu}_{i,h}-c_q\widehat{\sigma}_{i,h},
        \widehat{\mu}_{i,h}+c_q\widehat{\sigma}_{i,h}
      \right],
    \qquad
    c_q=\Phi^{-1}\!\left(\frac{1+q}{2}\right),
    \label{eq:prediction_interval}
\end{equation}
where $\Phi$ is the standard normal cumulative distribution function.
Prediction interval coverage probability (PICP) is the observed
fraction of targets within these intervals:
\begin{equation}
    \mathrm{PICP}(q)
    = \frac{1}{N_{\mathrm{test}}H}
      \sum_{i=1}^{N_{\mathrm{test}}}\sum_{h=1}^{H}
      \mathbf{1}\!\left\{y_{i,h}\in I_{i,h}(q)\right\}.
    \label{eq:picp}
\end{equation}
We report PICP@90, corresponding to $q=0.9$.
Calibration at this level is assessed by proximity to $90\%$.

Mean prediction interval width (MPIW) measures the average width
of these intervals. At the $90\%$ level,
\begin{equation}
    \mathrm{MPIW}@90
    = \frac{1}{N_{\mathrm{test}}H}
      \sum_{i=1}^{N_{\mathrm{test}}}\sum_{h=1}^{H}
      2c_{0.9}\widehat{\sigma}_{i,h}.
    \label{eq:mpiw90}
\end{equation}
We assess MPIW@90 alongside PICP@90: narrow intervals are useful
when their coverage remains close to the nominal level.

Our regression ECE averages the absolute coverage error over
$\mathcal{Q}=\{0.1,0.2,\ldots,0.9\}$:
\begin{equation}
    \mathrm{ECE}
    = \frac{1}{|\mathcal{Q}|}
      \sum_{q\in\mathcal{Q}}\left|\mathrm{PICP}(q)-q\right|.
    \label{eq:regression_ece}
\end{equation}
Lower ECE indicates closer agreement between nominal and empirical
coverage across these levels.

\paragraph{Distribution shift.}
For downstream forecasts on incoming datasets, we report prediction
interval coverage, width, and calibration using PICP@90, MPIW@90,
and ECE.
To assess the response to shift severity, we compute Spearman
correlations between measured dataset distance and the uncertainty
ratio, separately for each encoder and experiment.
The distance definition and evaluation panels are given in
\S\ref{subsec:distribution_shift}.

\paragraph{Aggregation across runs.}
For each incoming dataset, forecasting results report the mean and
sample standard deviation over 3 seeds $40$, $41$, and $42$.
Overall means give equal weight to the seven dataset means.
Routing comparisons pair runs by dataset and seed.
For the family coverage-error summary, we first average coverage across seeds within each dataset. We then compute the absolute deviation from $90\%$ and average these deviations across datasets.
 \section{Parameter and storage accounting}
\label{app:posterior_storage}

\begin{figure}[tbp]
\centering
\includegraphics[width=0.85\linewidth]{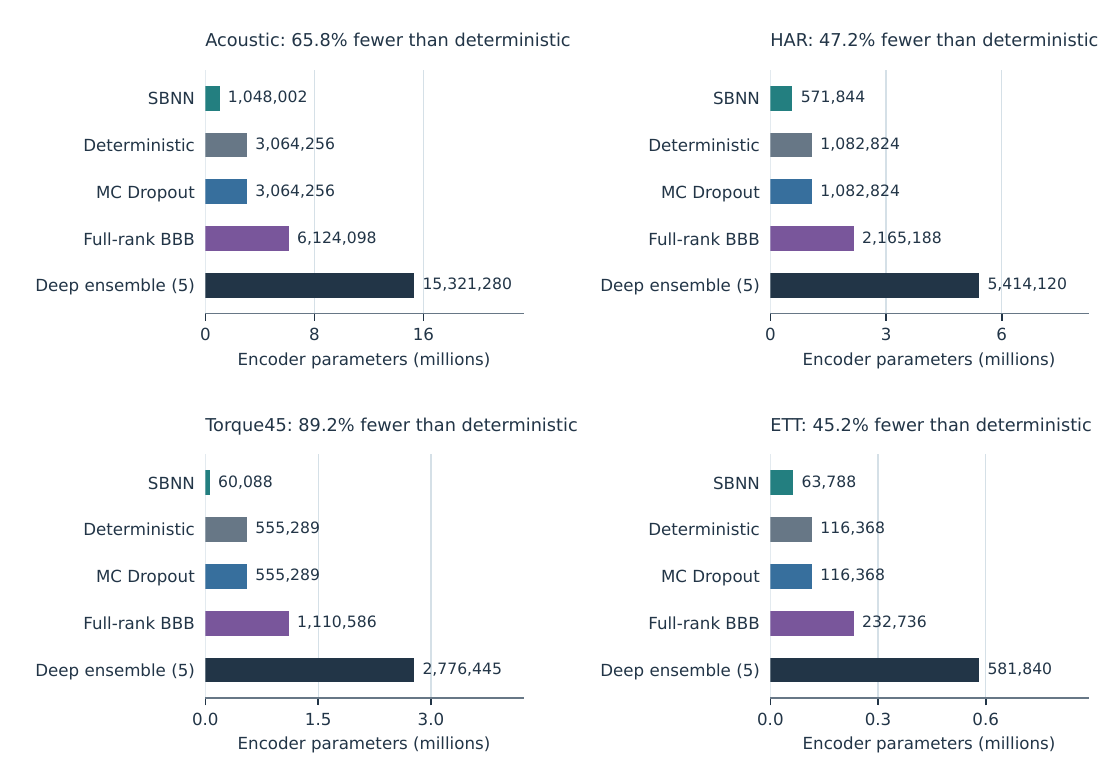}
\caption{\textbf{Stored encoder parameters by source and family.}
Percentages give SBNN reductions relative to deterministic encoders.
Panels use separate scales. Decoders and forecasting heads are excluded.}
\label{fig:source_encoder_storage}
\end{figure}

The source-specific encoder counts are reported in
Table~\ref{tab:encoder_storage}.

\paragraph{Forecasting head and pipeline.}
Table~\ref{tab:family_pipeline_parameter_breakdown}
separates the 17-encoder collection totals from the forecasting
head counts used in the storage comparison.
All heads use the same architecture and predict both the forecast and its conditional standard deviation. Their stored parameter counts differ because of their weight parameterizations.

\begin{table}[H]
\centering
\small
\setlength{\tabcolsep}{6pt}
\renewcommand{\arraystretch}{1.12}
\caption{\textbf{Encoder and forecasting-head parameter counts.}
Encoder totals cover the full 17-encoder collection in each family.
The total includes the collection and forecasting head.
Bayesian counts include posterior means and scales;
ensemble counts include all five members.
Bold indicates the smallest count in each column.}
\label{tab:family_pipeline_parameter_breakdown}
\begin{tabular}{@{}lrrr@{}}
\hline
Family & Encoders & Head & Total \\
\hline
SBNN
& \textbf{1,743,722}
& \textbf{14,784}
& \textbf{1,758,506} \\
Deterministic
& 4,818,737 & 49,600 & 4,868,337 \\
MC Dropout
& 4,818,737 & 49,600 & 4,868,337 \\
Full-rank BBB
& 9,632,608 & 98,752 & 9,731,360 \\
Deep ensemble (5)
& 24,093,685 & 248,000 & 24,341,685 \\
\hline
\end{tabular}
\end{table}

\paragraph{Parameter accounting.}
We count stored scalar model parameters, including frozen learned
parameters. Bayesian counts include posterior means and scales.
Decoder parameters are counted separately and excluded from forecasting. Fixed projection matrices and training-set standardization statistics are preprocessing state and are excluded from these counts. Their storage is separate from the reported model-parameter storage.
Optimizer state and metric accumulators are excluded. 

Let $p_{m,f}$ be the parameter count of encoder $m$ in family $f$,
and let $\mathcal{S}_f$ denote that family's selected set of $K$ experts.
The deployed pipeline count is
\begin{equation}
    P_f(\mathcal{S}_f)
    = \sum_{m\in\mathcal{S}_f}p_{m,f}+p_{\mathrm{head},f}.
    \label{eq:pipeline_parameter_count}
\end{equation}
Here, $p_{\mathrm{head},f}$ counts the deployed forecasting head
parameters. Ensemble encoder and head counts include all five members.
Each uncertainty family selects its own experts; the deterministic
pipeline reuses the SBNN selection (\S\ref{subsec:baselines_protocol}).
Collection storage sums the corresponding encoder costs over every
expert in the collection, regardless of routing selections.
These collection totals compare matched counterparts across families.
At the same numerical precision, parameter-storage ratios equal
parameter-count ratios.

The encoder-only reduction relative to the deterministic
counterparts is 63.8\%.
Including the forecasting head gives reductions of
63.9\% relative to the deterministic and MC Dropout models,
81.9\% relative to full-rank BBB, and 92.8\% relative
to the five-member ensemble.
\par A rank-matched deterministic factorization requires fewer stored parameters, but does not itself provide the epistemic uncertainty used by our router. Our objective is to obtain compact experts whose uncertainty also supports label-free selection. The reported storage reductions compare these Bayesian experts with their full-rank deterministic counterparts.\par \paragraph{Stored components of alternative Bayesian methods.}
Table~\ref{tab:bayesian_storage_components} summarizes what each
parameterization retains for prediction, including frozen weights.

\begin{table}[H]
\centering
\small
\setlength{\tabcolsep}{5pt}
\renewcommand{\arraystretch}{1.15}
\caption{\textbf{Stored components of Bayesian parameterizations.}
The comparison describes the published constructions, including
frozen backbone weights.}
\label{tab:bayesian_storage_components}
\begin{tabular}{@{}p{0.30\linewidth}p{0.66\linewidth}@{}}
\hline
Method & Stored components \\
\hline
SBNN~\citep{Toure2026}
& Means and scales of the low-rank weight factors; no retained dense
weight matrix for the factorized layer. \\
Rank-1 BNNs~\citep{Dusenberry2020}
& Shared dense weight matrix and posterior parameters for
multiplicative vectors. \\
SWAG~\citep{Maddox2019}
& Dense weight mean, diagonal variance statistics and retained
low-rank covariance deviation vectors. \\
Low-rank Gaussian VI~\citep{tomczak2020efficient}
& Dense weight mean and low-rank covariance factors, with a fixed
diagonal constant or a learned diagonal variance vector. \\
BLoB~\citep{wang2024blob}
& Frozen dense backbone, one deterministic adapter factor, and
the mean and scale of the other adapter factor. \\
\hline
\end{tabular}
\end{table}

\section{Routing protocol details}
\label{app:routing_protocol_details}

Routing is performed once for each incoming dataset.
The selected encoders and their aggregation weights then
remain fixed during head training and forecasting.
Neither selection method uses forecast targets.

\paragraph{Incoming inputs and source references.}
We compute routing scores from 192 unlabeled windows,
each containing 96 observations, sampled at evenly spaced
positions within the first 70\% of the incoming series.
These windows form a separate routing probe.
They are mapped to each encoder's input format using
the preprocessing described in
\S\ref{subsec:uncertainty_routing}.

For each encoder, the source uncertainty reference is
computed with the statistic in
Equation~\ref{eq:latent_epistemic_uncertainty}.
The incoming uncertainty is divided by this stored
reference, as defined in
Equation~\ref{eq:source_normalized_uncertainty}.
Only this scalar reference is retained for routing.

Each source reference is computed from 192 windows
from the encoder's own source dataset.
We use 40 posterior weight draws to estimate both
the source reference and the incoming routing uncertainty.

\paragraph{Uncertainty-based selection.}
Selection and weighting follow Table~\ref{tab:selection_policies}.

The nine-policy comparison uses a pool of 17 encoders.
Its low-uncertainty shortlist contains the eight encoders
with the lowest uncertainty ratios.

\paragraph{Source-distance selection.}
The distance baseline uses the statistical-feature
distance defined in
Equation~\ref{eq:source_distance_score}.
All encoder variants trained on the same source receive
that source's distance score.
The $K$ lowest-scoring candidates are selected and receive
normalized inverse-distance weights.

When several candidates have the same distance, their
selection follows a fixed ordering of encoder variants.
The selected set can therefore contain several encoders
from the same source. Candidates with equal distance
receive equal weights.

This baseline tests source-distance selection.
It does not reproduce the full two-stage routing method
of \citet{Pham2026Modular}.

\paragraph{Candidate eligibility.}
The routing comparisons use a pool of 17 encoders.

The distance-versus-uncertainty experiment compares both
selection and weighting.
Its results therefore assess the two routing policies
as a whole.

\section{Additional source reconstruction results}
\label{app:source_results}

\begin{table}[!htbp]
\centering
\footnotesize
\caption{\footnotesize\textbf{Source reconstruction and calibration.}
MSE uses validation windows; uncertainty metrics follow each source's
protocol described below. Bold marks the lowest
MSE or ECE, or coverage closest to 90\%. Dashes denote unreported values.}
\label{tab:source_reconstruction_calibration}
\setlength{\tabcolsep}{3pt}
\renewcommand{\arraystretch}{1.05}
\begin{minipage}[t]{0.49\linewidth}
\vspace{0pt}
\centering
\begin{tabular}{@{}lrrrr@{}}
\hline
\multicolumn{5}{l}{\textit{Acoustic}} \\
Model & MSE $\downarrow$ & ECE $\downarrow$
& \shortstack{PICP90 (\%)\\$\rightarrow 90\%$} & MPIW90 \\
\hline
Deterministic & 0.07613 & --- & --- & --- \\
MC Dropout & 0.07613 & 0.3620 & 29.48 & 0.2880 \\
Deep ensemble (5) & \textbf{0.05515} & 0.3391 & 30.86 & 0.2865 \\
Full-rank BBB & 0.07492 & 0.1251 & 71.09 & 0.8453 \\
SBNN & 0.10601 & \textbf{0.0453} & \textbf{80.72} & 1.3125 \\
\hline
\end{tabular}
\end{minipage}\hfill
\begin{minipage}[t]{0.49\linewidth}
\vspace{0pt}
\centering
\begin{tabular}{@{}lrrrr@{}}
\hline
\multicolumn{5}{l}{\textit{HAR}} \\
Model & MSE $\downarrow$ & ECE $\downarrow$
& \shortstack{PICP90 (\%)\\$\rightarrow 90\%$} & MPIW90 \\
\hline
Deterministic & 0.08142 & --- & --- & --- \\
MC Dropout & 0.08189 & 0.4301 & 14.86 & 0.0491 \\
Deep ensemble (5) & \textbf{0.08064} & \textbf{0.0602} & 72.46 & 0.2012 \\
Full-rank BBB & 0.08076 & 0.0686 & \textbf{78.27} & 0.2816 \\
SBNN & 0.10005 & 0.1114 & 77.62 & 0.3291 \\
\hline
\end{tabular}
\end{minipage}
\par\smallskip
\begin{minipage}[t]{0.49\linewidth}
\vspace{0pt}
\centering
\begin{tabular}{@{}lrrrr@{}}
\hline
\multicolumn{5}{l}{\textit{Torque45}} \\
Model & MSE $\downarrow$ & ECE $\downarrow$
& \shortstack{PICP90 (\%)\\$\rightarrow 90\%$} & MPIW90 \\
\hline
Deterministic & 0.03571 & --- & --- & --- \\
MC Dropout & 0.15339 & 0.1520 & 66.76 & 0.5721 \\
Deep ensemble (5) & \textbf{0.03551} & 0.4396 & 13.22 & 0.0660 \\
Full-rank BBB & 0.03579 & 0.3287 & 37.08 & 0.1842 \\
SBNN & 0.03594 & \textbf{0.0093} & \textbf{88.60} & 0.6147 \\
\hline
\end{tabular}
\end{minipage}\hfill
\begin{minipage}[t]{0.49\linewidth}
\vspace{0pt}
\centering
\begin{tabular}{@{}lrrrr@{}}
\hline
\multicolumn{5}{l}{\textit{ETT}} \\
Model & MSE $\downarrow$ & ECE $\downarrow$
& \shortstack{PICP90 (\%)\\$\rightarrow 90\%$} & MPIW90 \\
\hline
Deterministic & 0.09003 & --- & --- & --- \\
MC Dropout & 0.11859 & 0.2836 & 44.92 & 0.4341 \\
Deep ensemble (5) & 0.08458 & 0.3951 & 19.79 & 0.1800 \\
Full-rank BBB & \textbf{0.08257} & 0.4422 & 12.58 & 0.0867 \\
SBNN & 0.09567 & \textbf{0.1435} & \textbf{72.49} & 0.8026 \\
\hline
\end{tabular}
\end{minipage}
\end{table}

\paragraph{Reconstruction accuracy and calibration.}
Table~\ref{tab:source_reconstruction_calibration} reports reconstruction
accuracy and uncertainty calibration on the source domains, before
routing or downstream forecasting. The evaluated configurations
and additional SBNN selection roles are given in
Table~\ref{tab:source_all_configurations}.

For source reconstruction, Acoustic and ETT use empirical quantiles
of sampled reconstructions to form prediction intervals.
HAR and Torque45 use Gaussian intervals computed from the sampled
reconstruction means and standard deviations.
Calibration uses masked positions for Acoustic and ETT and the
full target windows for HAR and Torque45.

\begin{figure}[!htbp]
\centering
\includegraphics[width=\linewidth]{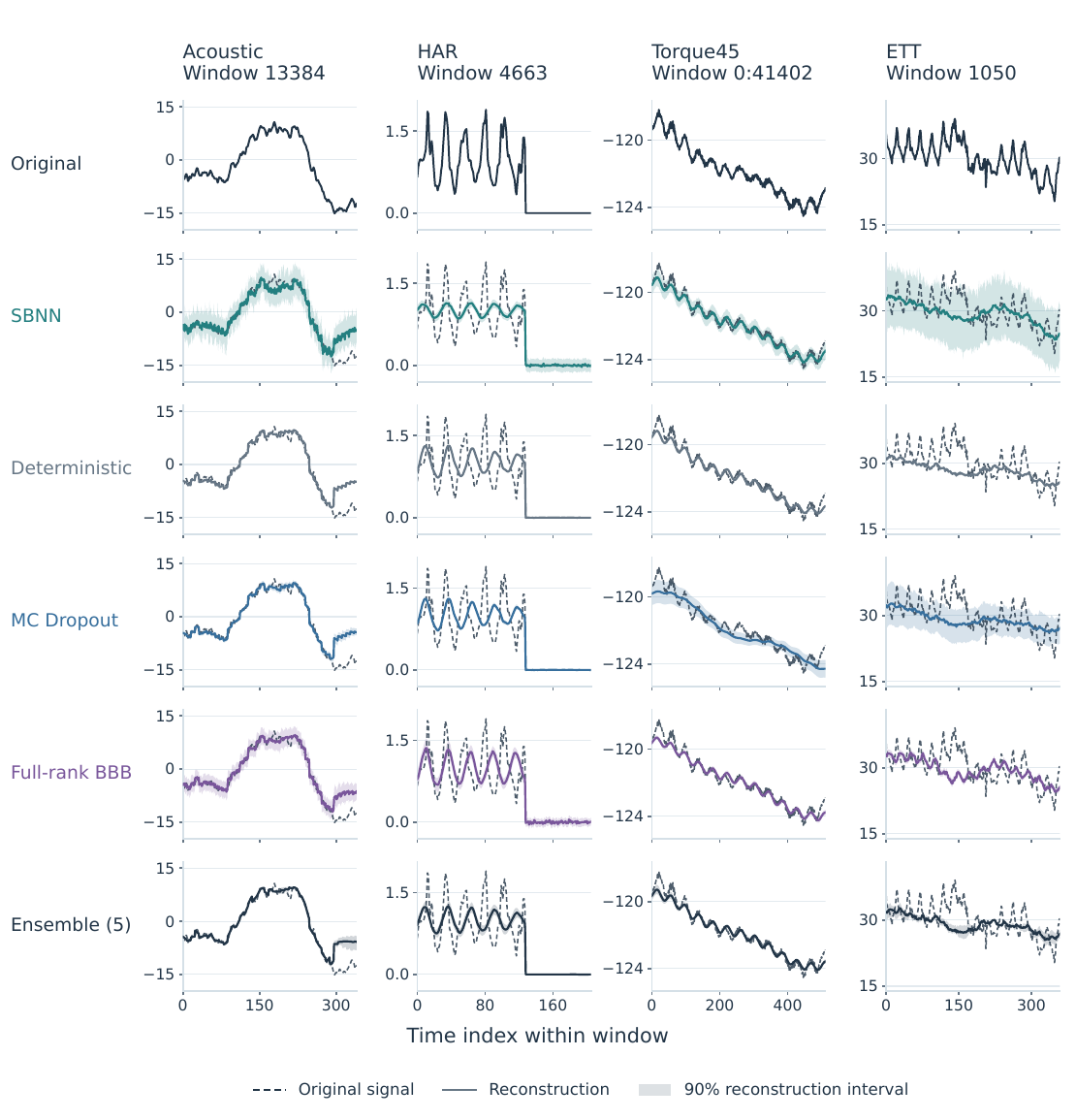}
\caption{\textbf{Source reconstructions and uncertainty bands.}
Selected validation windows from the four source domains.
The first row shows the original signal. Subsequent rows show
reconstructions (solid), the original signal (dashed) and nominal
90\% uncertainty bands. Models share the same window and axes
within each column. The deterministic model has no uncertainty band.}
\label{fig:source_reconstruction_comparison}
\end{figure}

Figure~\ref{fig:source_reconstruction_comparison} compares source reconstructions
and uncertainty bands across model families.

The selected SBNN encoders achieve the lowest ECE and coverage closest
to the nominal 90\% level on three of the four source domains.
HAR is the exception; its data-quality note is given in
Appendix~\ref{app:har_data_quality}.
SBNN does not achieve the lowest reconstruction MSE.
Its advantage is better uncertainty calibration at a lower storage
cost, with a source-dependent increase in reconstruction error.
The wider SBNN intervals cover more of the signal, while several
baselines give narrow intervals that miss much of it despite low MSE.

This accuracy-calibration tradeoff matters for expert selection.
Low reconstruction MSE can coexist with overconfident uncertainty
estimates. We therefore prioritize uncertainty quality alongside
reconstruction accuracy. The following shift and routing experiments
evaluate how the encoders' epistemic uncertainty supports expert
selection.
Rank and KL scale are tuned for compression with competitive
reconstruction and calibration (Appendix~\ref{app:rank_kl_sweep}).

The table reports the source configurations evaluated in this
reconstruction comparison. 

\subsection{Source results across SBNN configurations}

Table~\ref{tab:source_all_configurations} extends the
source-domain comparison in
Table~\ref{tab:source_reconstruction_calibration}.
These results concern reconstruction on the source datasets,
before routing or downstream forecasting.

For Acoustic, HAR and Torque45, we report three SBNN
selection priorities.
The reconstruction-selected configuration prioritizes
reconstruction accuracy, the calibration-selected configuration
prioritizes uncertainty quality, and the balanced configuration
considers both.
ETT reports its retained rank-KL configuration.

\begin{table}[H]
\centering
\small
\setlength{\tabcolsep}{5pt}
\renewcommand{\arraystretch}{1.10}
\caption{\textbf{Source reconstruction results across model
families and SBNN configurations.}
Bold indicates the lowest MSE or ECE, or coverage closest
to 90\%, within each source.
Dashes indicate unreported uncertainty metrics.}
\label{tab:source_all_configurations}
\begin{tabular}{@{}lrrrr@{}}
\hline
Model
& MSE $\downarrow$
& ECE $\downarrow$
& \shortstack{PICP90 (\%)\\$\rightarrow 90\%$}
& MPIW90 \\
\hline
\multicolumn{5}{l}{\textit{Acoustic}} \\
Deterministic
& 0.07613 & --- & --- & --- \\
MC Dropout
& 0.07613 & 0.3620 & 29.48 & 0.2880 \\
Deep ensemble (5)
& \textbf{0.05515} & 0.3391 & 30.86 & 0.2865 \\
Full-rank BBB
& 0.07492 & 0.1251 & 71.09 & 0.8453 \\
SBNN (reconstruction)
& 0.09097 & 0.0898 & 74.79 & 1.0428 \\
SBNN (calibration)
& 0.10601 & \textbf{0.0453} & \textbf{80.72} & 1.3125 \\
SBNN (balanced)
& 0.10164 & 0.0572 & 79.10 & 1.2267 \\
\hline
\multicolumn{5}{l}{\textit{HAR}} \\
Deterministic
& 0.08142 & --- & --- & --- \\
MC Dropout
& 0.08189 & 0.4301 & 14.86 & 0.0491 \\
Deep ensemble (5)
& \textbf{0.08064} & \textbf{0.0602} & 72.46 & 0.2012 \\
Full-rank BBB
& 0.08076 & 0.0686 & \textbf{78.27} & 0.2816 \\
SBNN (reconstruction)
& 0.09331 & 0.1189 & 76.56 & 0.3205 \\
SBNN (calibration)
& 0.10005 & 0.1114 & 77.62 & 0.3291 \\
SBNN (balanced)
& 0.10284 & 0.1141 & 73.10 & 0.2967 \\
\hline
\multicolumn{5}{l}{\textit{Torque45}} \\
Deterministic
& 0.03571 & --- & --- & --- \\
MC Dropout
& 0.15339 & 0.1520 & 66.76 & 0.5721 \\
Deep ensemble (5)
& \textbf{0.03551} & 0.4396 & 13.22 & 0.0660 \\
Full-rank BBB
& 0.03579 & 0.3287 & 37.08 & 0.1842 \\
SBNN (reconstruction)
& 0.03594 & 0.1038 & 77.49 & 0.4604 \\
SBNN (calibration)
& 0.03594 & \textbf{0.0093} & \textbf{88.60} & 0.6147 \\
SBNN (balanced)
& 0.03594 & 0.0647 & 94.39 & 0.7698 \\
\hline
\multicolumn{5}{l}{\textit{ETT}} \\
Deterministic
& 0.09003 & --- & --- & --- \\
MC Dropout
& 0.11859 & 0.2836 & 44.92 & 0.4341 \\
Deep ensemble (5)
& 0.08458 & 0.3951 & 19.79 & 0.1800 \\
Full-rank BBB
& \textbf{0.08257} & 0.4422 & 12.58 & 0.0867 \\
SBNN (retained)
& 0.09567 & \textbf{0.1435} & \textbf{72.49} & 0.8026 \\
\hline
\end{tabular}
\end{table}

The additional configurations show how the selection priority
changes the accuracy-calibration balance.
On Acoustic, prioritizing calibration lowers ECE and improves
coverage at the cost of higher reconstruction MSE.
On Torque45, the three SBNN configurations have the same MSE
to the displayed precision, but different interval widths
and coverage. The calibration-selected configuration is
closest to nominal coverage.
On HAR, the ensemble and BBB retain stronger calibration
results than the reported SBNN configurations.
\subsection{Source reconstruction examples}

The following figures compare original signals, reconstructions
and uncertainty intervals on selected source windows.

\begin{figure}[H]
\centering
\includegraphics[width=\linewidth]
{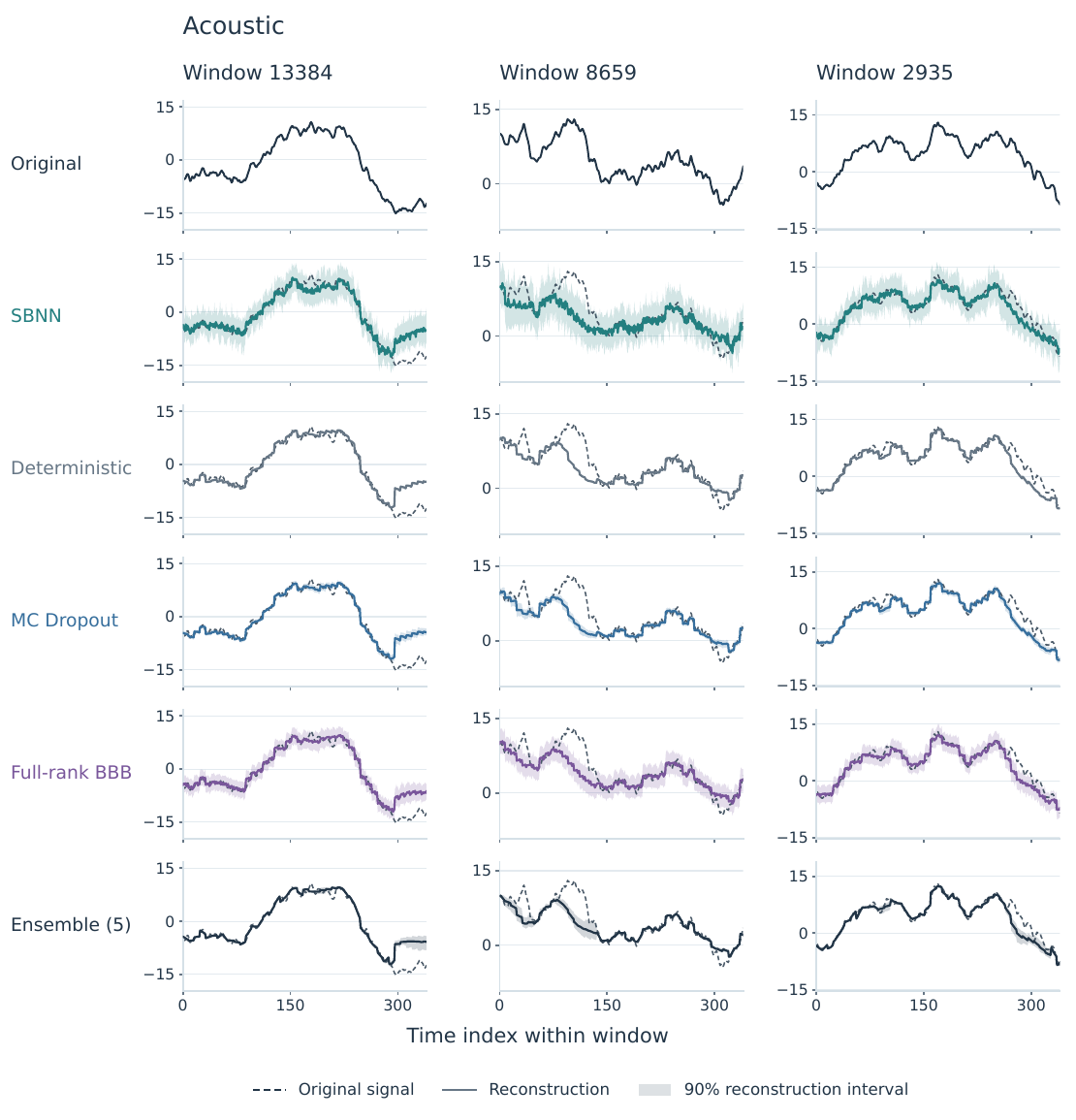}
\caption{\textbf{Acoustic source reconstruction.}
Each column shows the same window across all model families.
The first row displays the original signal.
Subsequent rows show reconstructions with nominal 90\%
uncertainty bands; dashed curves indicate the original signal.
The deterministic model has no uncertainty band.}
\label{fig:source_reconstruction_detail_acoustic}
\end{figure}
\begin{figure}[H]
\centering
\includegraphics[width=\linewidth]
{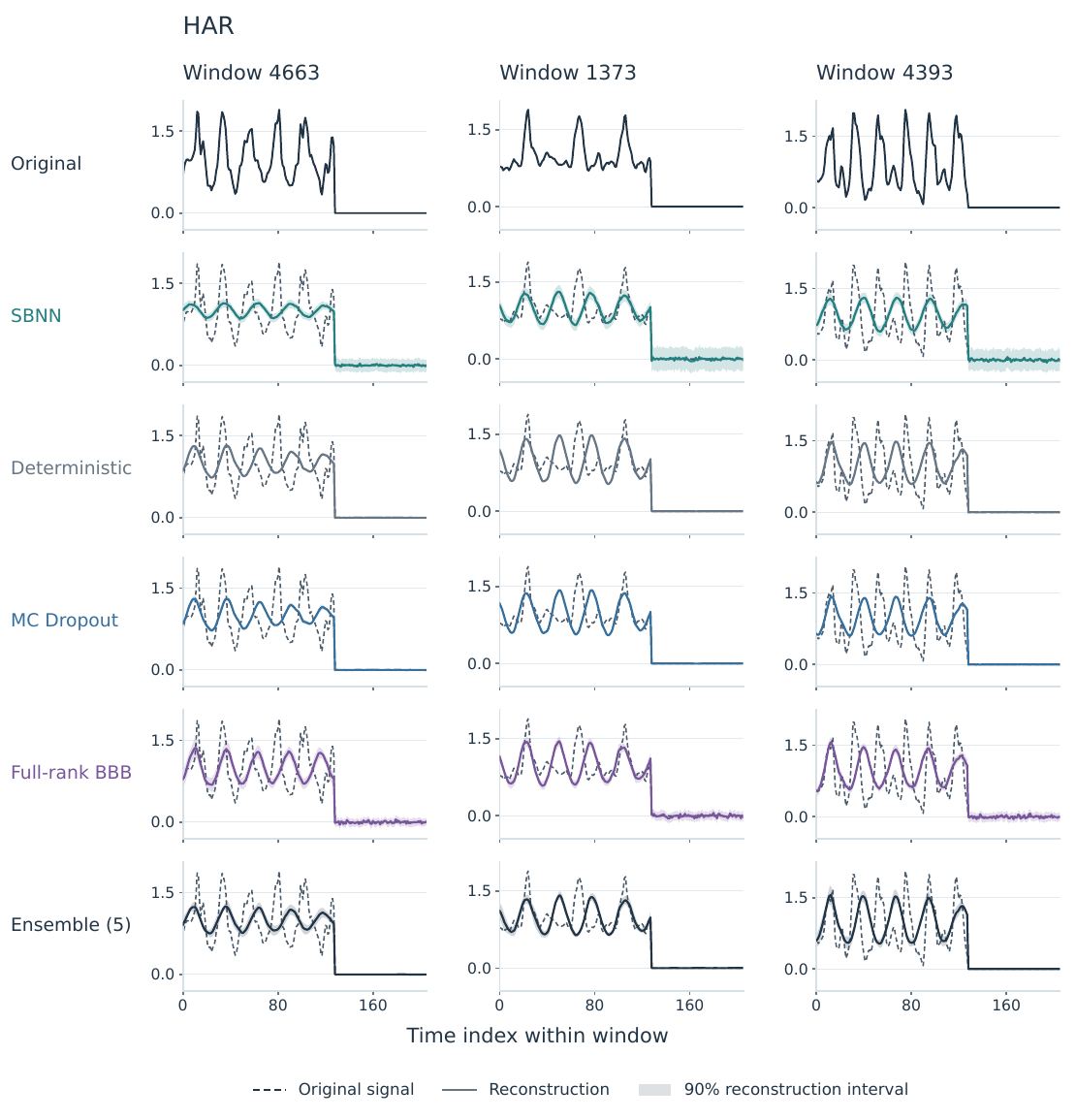}
\caption{\textbf{HAR source reconstruction.}
Each column shows the same window across all model families.
The first row displays the original signal.
Subsequent rows show reconstructions with nominal 90\%
uncertainty bands; dashed curves indicate the original signal.
The deterministic model has no uncertainty band.}
\label{fig:source_reconstruction_detail_har}
\end{figure}
\subsection{HAR data quality}
\label{app:har_data_quality}

We use the supplied HAR windows for signal reconstruction.
Activity labels do not enter the pretext objective.
The data preparation removes 26 windows identified by an
outlier filter before selecting the source-training subset.
All model families use the same prepared inputs.

The displayed windows contain varying-amplitude, pulse-like
patterns followed by flat segments at zero.
Their structure should be considered when interpreting
reconstruction errors and prediction intervals.

On this source, the deep ensemble achieves the lowest ECE,
while full-rank BBB has coverage closest to the nominal
90\% level.

\section{Additional distribution-shift results}
\label{app:shift_protocol}

This section provides the values supporting the
distribution-shift analysis in
\S\ref{subsec:distribution_shift}.

\subsection{Statistical-feature distance}

For each dataset, we construct a statistical descriptor
from 192 input windows of length 96.
It contains the mean, population standard deviation,
10th, 50th and 90th percentiles, and autocorrelations
at lags 1, 5 and 10:
\begin{equation}
    \phi(\mathcal{X})
    =
    \bigl[
    \mu,\sigma,q_{0.10},q_{0.50},q_{0.90},
    \mathrm{ACF}_{1},\mathrm{ACF}_{5},\mathrm{ACF}_{10}
    \bigr].
    \label{eq:shift_dataset_descriptor}
\end{equation}
The mean, standard deviation and quantiles are computed
over all values in the selected windows.
Autocorrelations are computed within each window and
then averaged across windows.
For each lag, we centre the window and divide the sum
of lagged products by its total centred sum of squares.

The distance follows Equation~\ref{eq:source_distance_score}.
Each coordinate scale $c_j$ is the population standard deviation
of that descriptor across Acoustic, HAR, Torque45 and ETT.
These scales remain fixed across comparisons.\par Descriptors use the stored window values without additional per-window standardization. The score therefore remains sensitive to differences in level and amplitude as well as temporal dependence.

\subsection{Distance and encoder uncertainty}

We evaluate one fixed encoder from each of the Torque45,
Acoustic and ETT sources across eight domains, including
each encoder's source.
For each encoder--dataset pair, we compute the uncertainty ratio
defined in \S\ref{subsec:uncertainty_routing} using 50 posterior draws.
No decoder or forecasting head is involved.
We compare the statistical descriptors of the source
and incoming datasets using the distance defined in
Equation~\ref{eq:source_distance_score}.
Distance is used here to assess the uncertainty response.
Uncertainty-based routing uses the epistemic ratios.

Figure~\ref{fig:encoder_shift_response} shows a positive
association between statistical-feature distance and encoder
uncertainty. For Torque45 and ETT, uncertainty increases
with distance overall, with local reversals in the ordering.
Acoustic saturates faster: its uncertainty ratio changes
little between Torque10 and Torque90 and decreases slightly
on ECL despite the larger distance.
Spearman correlations across the eight domains are
0.929 for Torque45, 0.810 for Acoustic and 0.905 for ETT.

Table~\ref{tab:shift_distance_uncertainty} reports the
statistical-feature distance and source-normalized
uncertainty ratio for each evaluated domain.
The ratio follows
Equation~\ref{eq:source_normalized_uncertainty}.

\begin{table}[H]
\centering
\small
\setlength{\tabcolsep}{4pt}
\renewcommand{\arraystretch}{1.12}
\caption{\textbf{Statistical-feature distance and encoder
uncertainty across eight domains.}
$D$ is the distance from the encoder's source dataset;
$r$ is its source-normalized uncertainty ratio.}
\label{tab:shift_distance_uncertainty}
\begin{tabular}{@{}lrrrrrr@{}}
\hline
& \multicolumn{2}{c}{Torque45 encoder}
& \multicolumn{2}{c}{Acoustic encoder}
& \multicolumn{2}{c}{ETT encoder} \\
Dataset & $D$ & $r$ & $D$ & $r$ & $D$ & $r$ \\
\hline
Torque45
& 0.000 & 1.000
& 1.055 & 0.751
& 1.739 & 1.538 \\
Acoustic
& 1.055 & 3.177
& 0.000 & 1.000
& 1.939 & 2.000 \\
HAR
& 1.059 & 0.962
& 1.874 & 0.780
& 1.863 & 1.437 \\
ETT
& 1.739 & 7.759
& 1.939 & 1.982
& 0.000 & 1.000 \\
WTH
& 1.130 & 3.913
& 1.063 & 1.289
& 1.547 & 1.627 \\
Torque10
& 4.035 & 18.648
& 3.699 & 2.976
& 5.397 & 5.749 \\
Torque90
& 30.820 & 143.833
& 30.453 & 3.038
& 32.172 & 34.601 \\
ECL
& 419.565 & 1878.404
& 419.808 & 2.756
& 418.055 & 407.874 \\
\hline
\end{tabular}
\end{table}

For the Acoustic encoder, Torque45 and HAR produce
uncertainty ratios below one.
These values indicate less variation across posterior
representations than on Acoustic inputs.
Their statistical-feature distances remain positive;
the Acoustic source has distance zero.

\subsection{Why uncertainty ratios can fall below one} \label{app:subunit_uncertainty}  The source reference compares the dispersion of representations under a fixed weight posterior. The same posterior can produce more stable representations on a new dataset than on its source. For example, the Acoustic ratios below one in Table~\ref{tab:shift_distance_uncertainty} indicate lower posterior representation dispersion on those inputs. The following construction establishes that this behaviour is possible within the SBNN variational family.\par  For a fixed posterior $q$ and a finite input set $\mathcal{X}$, define the population counterpart of Equation~\ref{eq:latent_epistemic_uncertainty} as \[ \mathcal{U}_q(\mathcal{X}) =\frac{1}{|\mathcal{X}|d} \sum_{x\in\mathcal{X}}\sum_{j=1}^{d} \sqrt{\operatorname{Var}_{q}[z_j(x)]}, \] where $d$ is the representation dimension. With independent posterior draws and finite second moments, the sample-based statistic converges almost surely to this quantity as the number of draws increases.  \paragraph{A rank-one SBNN construction.} Consider $z(x)=AB^\top x$, with $A=(a_1,a_2)^\top$ and $B=(b_1,b_2)^\top$. Let all four factor entries be independent, with $a_1,a_2,b_1\sim\mathcal{N}(1,1)$ and $b_2\sim\mathcal{N}(2,4)$; the second Gaussian argument is the variance. This is a mean-field Gaussian factor posterior with strictly positive variances, and $AB^\top$ has rank one almost surely. Let the source and incoming input distributions be uniform on $\mathcal{X}_{\mathrm{src}}=\{e_2,-e_2\}$ and $\mathcal{X}_{\mathrm{in}}=\{e_1,-e_1\}$, respectively, where $e_1=(1,0)^\top$ and $e_2=(0,1)^\top$. Their supports are disjoint, and every input has unit norm.\par  For independent scalar factors $a$ and $b$, \[ \operatorname{Var}(ab) =\bigl(\operatorname{Var}(a)+\mathbb{E}[a]^2\bigr)   \operatorname{Var}(b) +\operatorname{Var}(a)\mathbb{E}[b]^2. \] For each coordinate $j\in\{1,2\}$, this gives $\operatorname{Var}(a_jb_1)=3$ and $\operatorname{Var}(a_jb_2)=12$. Input sign changes leave these variances unchanged. Consequently, $\mathcal{U}_q(\mathcal{X}_{\mathrm{in}})=\sqrt{3}$ and $\mathcal{U}_q(\mathcal{X}_{\mathrm{src}})=\sqrt{12}$, so their ratio is exactly $1/2$. Since the source quantity is positive, the corresponding Monte Carlo ratio converges almost surely to $1/2$ as the posterior draw counts for both sets increase.\par  The construction uses equal-norm inputs and a linear encoder, so neither a change in input amplitude nor activation saturation is required for a ratio below one. The weight posterior stays fixed; its induced representation dispersion depends on the input. This establishes theoretical compatibility with the observed sub-unit ratios. Identifying the particular mechanism in the trained Acoustic encoder would require a separate analysis. The broader dependence of Bayesian uncertainty on function-space properties is discussed by \citet{dangelo2022outofdistributiondetectionbayesianneural}.  \section{Forecasting loss and temporal fidelity}
\label{app:forecasting_loss_transition}

\paragraph{Initial forecasting behaviour.}
Our initial forecasting head was trained with MSE and
returned only a point prediction.
Although aggregate errors were low, inspection of the
forecast windows showed overly smooth predictions.
They stayed near the signal's average level and missed
much of its variation.
We describe this behaviour as flattening toward the mean.
This observation motivated the change to a probabilistic
head trained with $\beta$-NLL.

\paragraph{Learning the forecast and its uncertainty.}
The final head predicts a value $\mu$ and a positive
standard deviation $\sigma$ for each forecast coordinate.
The learned $\sigma$ represents the conditional aleatoric
uncertainty.
For a scalar target $y$, the loss is
\begin{equation}
    \mathcal{L}_{\beta\text{-NLL}}(y,\mu,\sigma)
    =
    \operatorname{stopgrad}\!\left((\sigma^2)^\beta\right)
    \frac{1}{2}
    \left[
        \log\sigma^2 + \frac{(y-\mu)^2}{\sigma^2}
    \right],
    \qquad \beta=0.5.
    \label{eq:beta_nll}
\end{equation}
The loss is averaged over forecast coordinates and training examples.
The stop-gradient operation holds the multiplicative weight fixed
when differentiating the loss. This weight reduces the downweighting
of mean-fitting errors by large predicted variances under Gaussian NLL.

Under ordinary Gaussian NLL, the gradient with respect
to the point prediction is
\begin{equation}
    \frac{\partial\mathcal{L}_{\mathrm{NLL}}}{\partial\mu}
    =
    \frac{\mu-y}{\sigma^2}.
    \label{eq:gaussian_nll_mean_gradient}
\end{equation}
A large predicted variance therefore reduces the
contribution of an observation to fitting the prediction.
\citet{Seitzer2022PitfallsOfUncertainty} show how this
weighting can hinder learning in poorly fitted regions.

The detached variance weight in $\beta$-NLL changes
this gradient to
\begin{equation}
    \frac{\partial\mathcal{L}_{\beta\text{-NLL}}}{\partial\mu}
    =
    (\mu-y)(\sigma^2)^{\beta-1}.
    \label{eq:beta_nll_mean_gradient}
\end{equation}
For $\beta=0.5$, the gradient is $(\mu-y)/\sigma$.
Compared with Gaussian NLL, high predicted variance
suppresses the prediction-fitting gradient less strongly.

MSE has no variance-dependent weighting.
Our transition therefore changes both the training
objective and the head's outputs.
The final head learns an observation-noise estimate,
which contributes to the prediction intervals alongside
epistemic uncertainty.
Temporal fidelity is examined through forecast windows;
accuracy and uncertainty quality are evaluated separately
using MSE, ECE, coverage and Gaussian NLL.

\paragraph{Bayesian head objective.}
For the Bayesian head, the objective is
\begin{equation}
    \mathcal{J}_{\mathrm{head}}(q_\phi)
    =
    \frac{1}{N}\sum_{i=1}^{N}
    \mathbb{E}_{\omega\sim q_\phi}
    \left[
        \mathcal{L}_{\beta\text{-NLL}}
        \!\left(y_i,\mu_i(\omega),\sigma_i(\omega)\right)
    \right]
    +\lambda_{\mathrm{head}}\mathcal{K}_{\mathrm{head}}.
    \label{eq:head_training_objective}
\end{equation}
Head training uses a fixed-feature approximation: the expectation
is over head weights only, with encoder weights fixed at their
posterior means. Encoder uncertainty is propagated at prediction.
The encoder and head posteriors are fitted in separate stages.
The head uses a KL-regularized $\beta$-NLL surrogate objective.

The term $\mathcal{K}_{\mathrm{head}}$ sums the head's factor-space KL
terms, as in \S\ref{subsec:bayesian_encoder_learning}, and
$\lambda_{\mathrm{head}}$ controls their strength.

Sampling encoder weights at prediction can change the feature distribution received by the head. Through nonlinear transformations, this can affect both forecasts and predicted conditional scales. The reported calibration metrics evaluate the complete predictive distribution obtained with this procedure. The effect of encoder sampling was not isolated experimentally.\par  Let $Q$ denote the sampling distribution over the frozen encoder posteriors and the fitted head posterior. For each forecast coordinate $Y$, the Gaussian mixture defined by our prediction procedure satisfies the law of total variance: \[ \operatorname{Var}(Y\mid x) = \operatorname{Var}_{Q}[\mu(x)] + \mathbb{E}_{Q}[\sigma^2(x)]. \] This identity gives the variance of the constructed predictive distribution; its calibration is assessed empirically.\par  \subsection{Initial forecasting without a learned aleatoric output}
\label{app:initial_forecasting}

The initial forecasting heads were trained with MSE and
returned only a point prediction. Prediction intervals
used epistemic uncertainty, without a learned aleatoric
contribution.

In that setup, SBNN had substantially
lower mean ECE and coverage much closer to the nominal 90\%
level than MC Dropout and deep ensembles.
With $\beta$-NLL, the head also predicts an aleatoric standard
deviation $\sigma$. Including this contribution in the predictive
intervals improves coverage across these families and substantially
narrows the calibration gap, as shown in the final comparison.

Table~\ref{tab:initial_forecasting_mean} reports the
average results across the seven incoming datasets.
The deterministic model produced no uncertainty intervals
in this initial setup.

\begin{table}[H]
\centering
\small
\setlength{\tabcolsep}{6pt}
\renewcommand{\arraystretch}{1.12}
\caption{\textbf{Initial forecasting with epistemic-only intervals.}
Results are averaged across seven incoming datasets.
Heads were trained with MSE.
A dash indicates an unavailable interval metric.}
\label{tab:initial_forecasting_mean}
\begin{tabular}{@{}lrrrr@{}}
\hline
Family & MSE $\downarrow$ & ECE $\downarrow$
& \shortstack{PICP90 (\%)\\$\rightarrow 90\%$}
& MPIW90 \\
\hline
SBNN
& 0.456 & \textbf{0.088} & \textbf{80.60} & 1.874 \\
Deterministic
& 0.407 & --- & --- & --- \\
MC Dropout
& \textbf{0.392} & 0.341 & 32.77 & 0.363 \\
Deep ensemble (5)
& 0.394 & 0.385 & 24.61 & 0.283 \\
\hline
\end{tabular}
\end{table}

SBNN provides substantially better average calibration
and coverage in this initial comparison, at the cost
of higher forecasting MSE.
Its ECE is lowest on six of the seven datasets,
and its coverage is closest to 90\% on all seven.
The much narrower MC Dropout and ensemble intervals
cover fewer than one third of the targets on average.

Section~\ref{subsec:forecasting} reports the final forecasting results with $\beta$-NLL heads and learned aleatoric uncertainty.

\section{Detailed model-family forecasting results}
\label{app:family_forecasting_details}

\paragraph{Comparison objectives.}
We separate the comparison of model families from the comparison
of routing policies. The first compares the storage and predictive quality of complete family-specific pipelines, including their routing. The second tests how expert
selection affects downstream forecasting.

\paragraph{Matched model families.}
Our routing collection contains SBNN encoders. For comparison, each
SBNN expert is matched to a counterpart trained on the same source
dataset in the deterministic, MC Dropout, full-rank Bayes by Backprop,
and five-member deep ensemble families. Each uncertainty family performs its own routing and computes
its own aggregation weights. The deterministic pipeline uses
the SBNN selections and weights. Each family
uses its own encoders and a corresponding forecasting head.
We compare source reconstruction and downstream forecasting,
including predictive uncertainty across the full encoder--head pipeline.
The deterministic pipeline uses fixed weights.
Its heteroscedastic regression head predicts the target $y$
and an input-dependent standard deviation $\sigma$.

Table~\ref{tab:family_forecasting_results} summarizes forecasting accuracy and storage; Appendix~\ref{app:posterior_storage} gives the encoder and head counts separately.

\paragraph{Predictive calibration and coverage.}
The intervals and calibration metrics use the total predictive
variance in Equation~\ref{eq:total_predictive_variance}.
It combines variation across sampled forecasts with the average
squared standard deviation predicted by the head.
The learned $\sigma$ contributes to interval coverage in every
family, including the deterministic pipeline, whose weights
remain fixed. It can improve coverage even when the epistemic
component is small.

SBNN has the lowest mean ECE, 0.0687, compared with 0.0734 for
the ensemble, 0.0767 for MC Dropout and 0.0877 for BBB.
Its mean NLL is also lowest, although the difference from
MC Dropout is small: 0.814 versus 0.817.
PICP90 reaches 90.47\%, closest to the nominal 90\% level.
MC Dropout and the ensemble reach 87.78\% and 84.55\%,
while BBB reaches 92.29\%.

Coverage must be considered together with interval width.
SBNN's mean MPIW90 is 1.771, above MC Dropout's 1.647 and
the ensemble's 1.517, and below BBB's 2.100.
Its improved coverage comes with wider intervals than the
under-covering MC Dropout and ensemble models.
The deterministic pipeline has the narrowest mean intervals.
They cover 82.60\% of the targets.

Algorithm~\ref{alg:joint_prediction} summarizes prediction after
routing and head training. The selected encoders, routing weights
and alignment maps remain fixed; encoder and head weights are
sampled at each iteration.

\begin{algorithm}[H]
\caption{Monte Carlo prediction through the selected encoder--head pipeline}
\label{alg:joint_prediction}
\begin{algorithmic}[1]
\Require Input window $x$, selected set of encoders $\mathcal{S}$, weights
    $\{\alpha_m\}_{m\in\mathcal{S}}$, number of draws $N_{\mathrm{MC}}$
\Require Frozen encoder posteriors $\{q_m\}_{m\in\mathcal{S}}$,
    preprocessing maps $T_m$, alignment maps $P_m$
\Require Trained head $g$ with weight posterior $q_\phi$
\Ensure Predictive mean and epistemic, aleatoric, and total variances
\For{$t=1,\ldots,N_{\mathrm{MC}}$}
    \State $h^{(t)} \gets 0$
    \For{each $m\in\mathcal{S}$}
        \State Sample $\theta_m^{(t)}\sim q_m$
        \State $z_m^{(t)} \gets f_m(T_m(x);\theta_m^{(t)})$
        \State $h^{(t)} \gets h^{(t)}+\alpha_m P_m(z_m^{(t)})$
    \EndFor
    \State Sample $\omega^{(t)}\sim q_\phi$
    \State $(\mu^{(t)},\sigma^{(t)}) \gets g(h^{(t)};\omega^{(t)})$
\EndFor
\State $\widehat{\mu} \gets
    N_{\mathrm{MC}}^{-1}\sum_{t=1}^{N_{\mathrm{MC}}}\mu^{(t)}$
\State $\widehat{v}_{\mathrm{epi}} \gets
    N_{\mathrm{MC}}^{-1}\sum_{t=1}^{N_{\mathrm{MC}}}
    (\mu^{(t)}-\widehat{\mu})^2$
\State $\widehat{v}_{\mathrm{alea}} \gets
    N_{\mathrm{MC}}^{-1}\sum_{t=1}^{N_{\mathrm{MC}}}(\sigma^{(t)})^2$
\State $\widehat{v}_{\mathrm{pred}} \gets
    \widehat{v}_{\mathrm{epi}}+\widehat{v}_{\mathrm{alea}}$
\State \Return $\widehat{\mu}$, $\widehat{v}_{\mathrm{epi}}$,
    $\widehat{v}_{\mathrm{alea}}$, $\widehat{v}_{\mathrm{pred}}$
\Statex \textit{Forecast means and variances are computed coordinatewise.}
\end{algorithmic}
\end{algorithm}

\subsection{Forecast examples across incoming datasets}
\begin{figure}[H]
\centering
\includegraphics[width=\linewidth]
{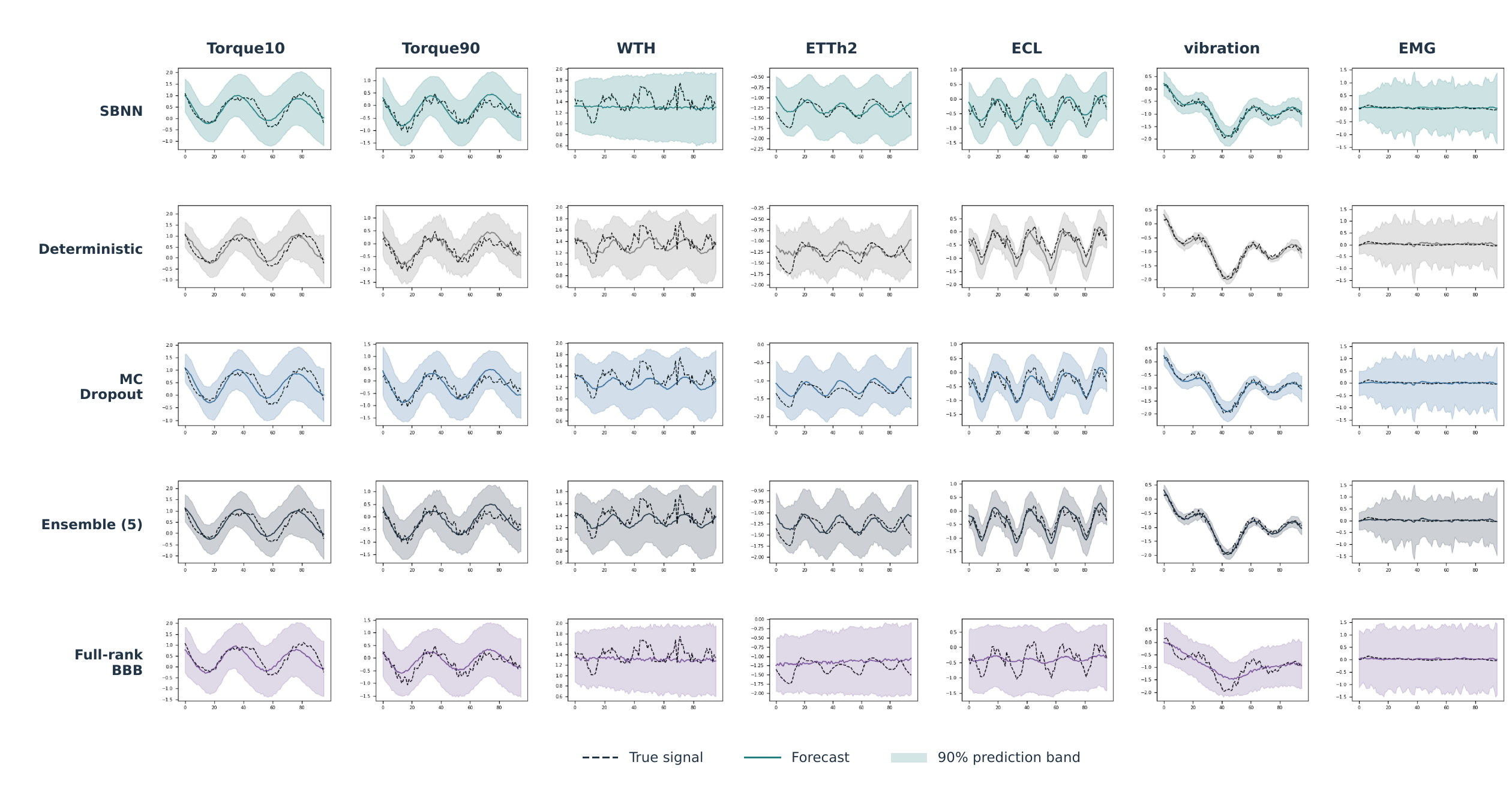}
\caption{\textbf{Forecasts and uncertainty bands across all seven incoming datasets.}
Columns show incoming datasets and rows show model families.
Each column uses the same saved forecast window across all five families.
Dashed curves show the true signal, solid curves the forecast,
and shaded regions nominal 90\% prediction bands.
Panels retain their original vertical scales.}
\label{fig:family_forecasting_windows}
\end{figure}

\subsection{Additional predictive-coverage comparisons}
\label{app:family_forecasting_figures}

SBNN has the lowest average absolute deviation from nominal
$90\%$ coverage: $3.15$ percentage points, compared with $3.32$
for full-rank BBB, $4.89$ for MC Dropout and $6.07$ for deep
ensembles. We compute this deviation from each dataset's
seed-averaged coverage before averaging across datasets, so
undercoverage and overcoverage do not cancel. Coverage varies across incoming datasets; near-nominal average coverage does not imply nominal coverage on every dataset. SBNN combines competitive forecasting and lower average coverage error with substantially fewer stored parameters.

\begin{figure}[H]
\centering
\includegraphics[width=\linewidth]
{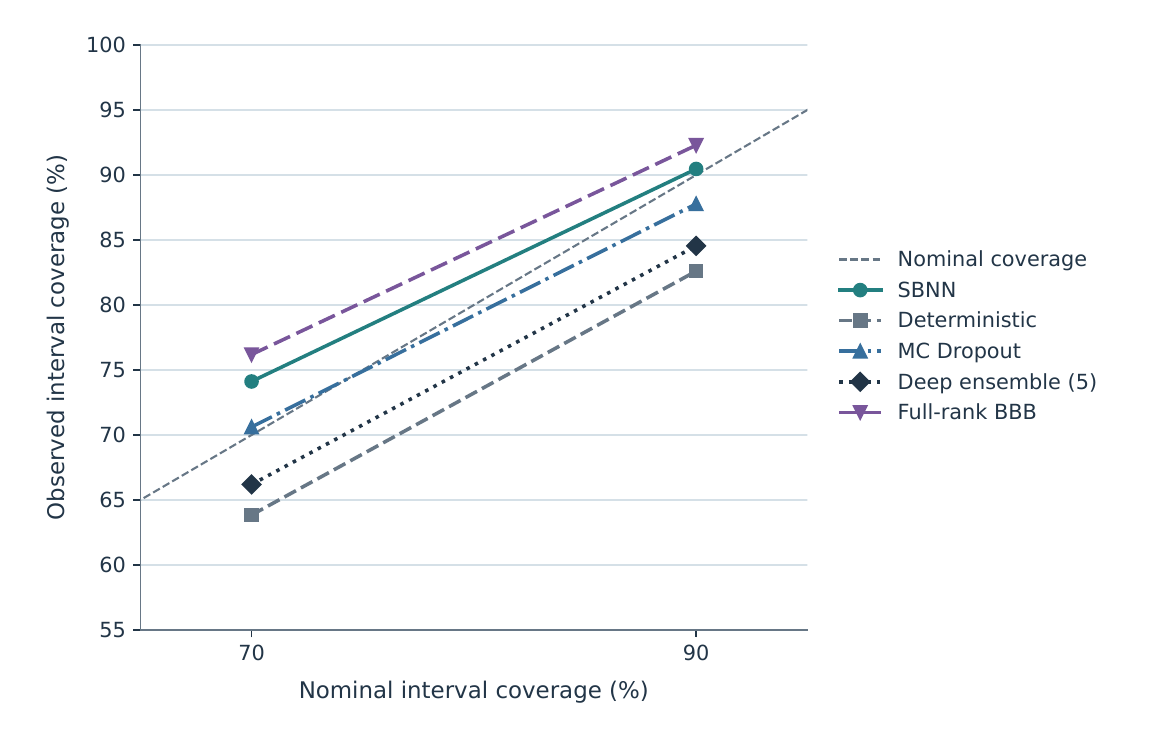}
\caption{\textbf{Forecasting coverage at two nominal levels.}
Points show observed coverage at nominal 70\% and 90\%,
averaged over seven incoming datasets and three seeds.
The dashed diagonal marks agreement between nominal
and observed coverage.}
\label{fig:family_coverage_profile}
\end{figure}

Figure~\ref{fig:family_coverage_profile} compares observed
coverage at nominal levels of 70\% and 90\%.
SBNN is closest to nominal coverage at 90\%,
while MC Dropout is closest at 70\%.

\subsection{Fixed projection and representation fusion} \label{app:projection_fusion}  For this derivation, index the selected experts by $m=1,\ldots,K$. Let $u_m\in\mathbb{R}^{d_m}$ denote encoder $m$'s standardized representation and $R_m\in\mathbb{R}^{64\times d_m}$ its fixed projection matrix, so $P_m(z_m)=R_m u_m$. The aggregation in Equation~\ref{eq:representation_aggregation} can be written as \[ h=R_{\alpha}u,\qquad R_{\alpha}=[\alpha_1R_1,\ldots,\alpha_KR_K],\qquad u=[u_1^\top,\ldots,u_K^\top]^\top. \] Thus, fusion is a fixed linear compression of the concatenated, standardized representations. It gives the head a constant input dimension across heterogeneous encoders and selected sets, without learning additional projection parameters. The head is fitted to this combined feature map; latent coordinates from different encoders need not have matching meanings. Compression can discard useful information. Comparisons with concatenation or learned projections remain future work.\par  \paragraph{Covariance of the fused representation.} Condition on the selected set, routing weights, standardization statistics and realized projection matrices. For a specified source of variation with finite second moments, write $C_{mn}=\operatorname{Cov}(u_m,u_n)$. Linearity gives the exact identity \[ \operatorname{Cov}(h) =\sum_{m=1}^{K}\sum_{n=1}^{K} \alpha_m\alpha_n R_m C_{mn}R_n^\top. \] Across input windows, the cross-covariances can be nonzero because all encoders process the same inputs. When variation comes only from independently sampled encoder weights at a fixed input $x$, $C_{mn}(x)=0$ for $m\ne n$. Writing $V_m(x)=R_m C_{mm}(x)R_m^\top$, the covariance then becomes \[ \operatorname{Cov}_{q}(h\mid x) =\sum_{m=1}^{K}\alpha_m^2 V_m(x). \] The weight identity $\sum_m\alpha_m^2\in[1/K,1]$ holds for nonnegative weights summing to one. If all projected expert covariances equal $V(x)$, the last expression reduces to $(\sum_m\alpha_m^2)V(x)$. Independence of the random matrix draws alone does not establish that equality for the realized projected expert covariances. Changing the selected experts, $K$ or their weights can therefore change feature scale. Each policy fits its forecasting head to its own fixed feature map.  \section{Detailed distance-versus-uncertainty routing results}
\label{app:routing_results}

\paragraph{Aggregate routing comparison.}
Table~\ref{tab:downstream_routing_comparison} summarizes the top-3 routing comparison in \S\ref{subsec:routing_comparison}.

\begin{table}[!htbp]
\centering
\vspace{0pt}
\centering
\footnotesize
\setlength{\tabcolsep}{3pt}
\renewcommand{\arraystretch}{1.10}
\caption{\footnotesize\textbf{Forecasting with top-3 distance and uncertainty routing.}
Mean $\pm$ sample SD over three seeds, each averaged over seven datasets.
Bold marks the better mean.}
\label{tab:downstream_routing_comparison}
\begin{tabular}{@{}lrr@{}}
\hline
Metric & \shortstack{Top-3\\distance} & \shortstack{Top-3\\uncertainty} \\
\hline
MSE $\downarrow$ & \(0.4062\pm0.0059\) & \(\mathbf{0.3976}\pm0.0070\) \\
ECE $\downarrow$ & \(0.0857\pm0.0013\) & \(\mathbf{0.0721}\pm0.0038\) \\
NLL $\downarrow$ & \(0.907\pm0.039\) & \(\mathbf{0.836}\pm0.041\) \\
\shortstack[l]{PICP90 (\%)\\$\rightarrow 90\%$}
& \(82.28\pm0.58\) & \(\mathbf{89.03}\pm0.35\) \\
MPIW90 & \(1.795\pm0.007\) & \(1.715\pm0.032\) \\
\hline
\end{tabular}
\end{table}

Table~\ref{tab:routing_results_by_dataset} gives the
per-dataset results for the comparison in
\S\ref{subsec:routing_comparison}.
Both policies select three SBNN encoders.
Entries report the mean and sample standard deviation
across seeds 40, 41 and 42.

\begin{figure}[H]
\centering
\includegraphics[width=\linewidth]
{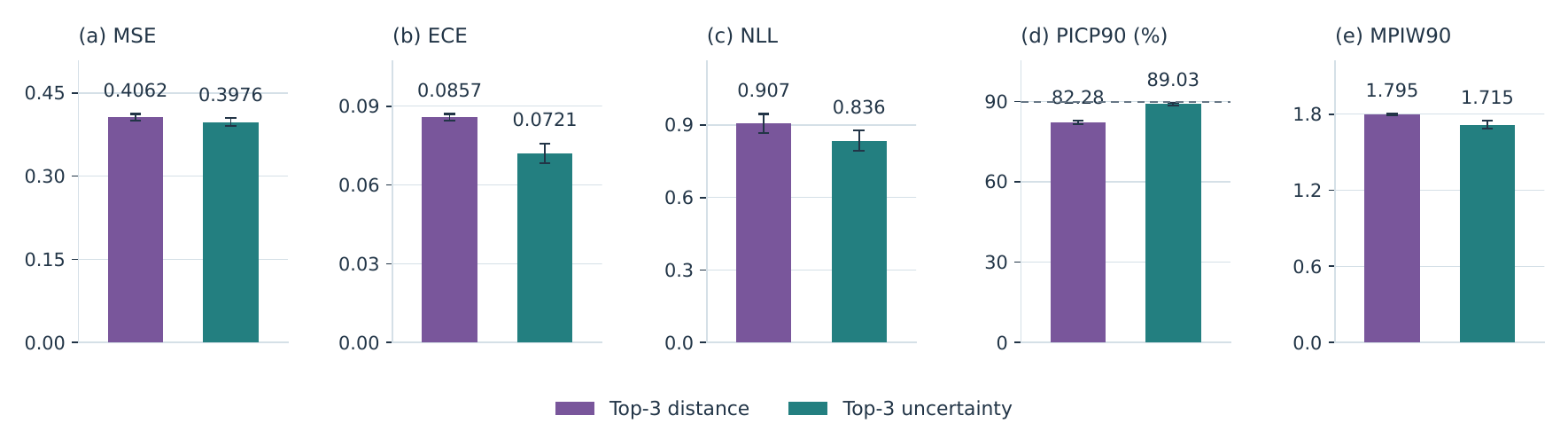}
\caption{\textbf{Forecasting under distance and uncertainty routing.}
Both policies select three SBNN encoders.
Bars show averages across seven incoming datasets.
Whiskers show sample standard deviations across three seeds,
after averaging each seed across datasets.
The dashed line marks nominal 90\% coverage.
}
\label{fig:routing_distance_uncertainty_additional_metrics}
\end{figure}

\begin{table}[H]
\centering
\footnotesize
\setlength{\tabcolsep}{3pt}
\renewcommand{\arraystretch}{1.1}
\caption{\textbf{Forecasting under distance and uncertainty routing.}
PICP90 is expressed as a percentage.
}
\label{tab:routing_results_by_dataset}
\begin{tabular}{@{}lrrrrr@{}}
\hline
Policy & MSE $\downarrow$ & ECE $\downarrow$ & NLL $\downarrow$
& \shortstack{PICP90 (\%)\\$\rightarrow 90\%$} & MPIW90 \\
\hline
\multicolumn{6}{l}{\textbf{Torque10}} \\
Top-3 distance
& $0.5531\pm0.0098$ & $0.0624\pm0.0086$
& $1.108\pm0.014$ & $82.62\pm1.81$ & $1.997\pm0.013$ \\
Top-3 uncertainty
& $0.5004\pm0.0159$ & $0.0419\pm0.0071$
& $1.067\pm0.020$ & $85.45\pm0.56$ & $2.038\pm0.010$ \\
\hline
\multicolumn{6}{l}{\textbf{Torque90}} \\
Top-3 distance
& $0.3164\pm0.0041$ & $0.0436\pm0.0036$
& $0.889\pm0.005$ & $83.52\pm0.22$ & $1.999\pm0.024$ \\
Top-3 uncertainty
& $0.3319\pm0.0029$ & $0.0195\pm0.0023$
& $0.808\pm0.001$ & $91.46\pm0.21$ & $1.914\pm0.010$ \\
\hline
\multicolumn{6}{l}{\textbf{WTH}} \\
Top-3 distance
& $0.3112\pm0.0026$ & $0.0142\pm0.0007$
& $0.800\pm0.001$ & $84.35\pm0.99$ & $1.615\pm0.017$ \\
Top-3 uncertainty
& $0.2869\pm0.0083$ & $0.0055\pm0.0015$
& $0.744\pm0.016$ & $88.74\pm0.33$ & $1.501\pm0.014$ \\
\hline
\multicolumn{6}{l}{\textbf{ETTh2}} \\
Top-3 distance
& $0.3219\pm0.0289$ & $0.0926\pm0.0010$
& $0.990\pm0.092$ & $74.39\pm0.82$ & $1.550\pm0.007$ \\
Top-3 uncertainty
& $0.3516\pm0.0366$ & $0.0874\pm0.0112$
& $0.951\pm0.071$ & $79.44\pm1.33$ & $1.452\pm0.017$ \\
\hline
\multicolumn{6}{l}{\textbf{ECL}} \\
Top-3 distance
& $0.4700\pm0.0091$ & $0.0096\pm0.0035$
& $1.009\pm0.002$ & $84.85\pm0.23$ & $2.036\pm0.029$ \\
Top-3 uncertainty
& $0.4708\pm0.0013$ & $0.0067\pm0.0022$
& $1.003\pm0.001$ & $88.69\pm0.21$ & $1.971\pm0.013$ \\
\hline
\multicolumn{6}{l}{\textbf{Vibration}} \\
Top-3 distance
& $0.1060\pm0.0007$ & $0.1580\pm0.0054$
& $0.137\pm0.004$ & $81.94\pm0.42$ & $1.214\pm0.008$ \\
Top-3 uncertainty
& $0.0853\pm0.0002$ & $0.1378\pm0.0066$
& $-0.067\pm0.004$ & $97.33\pm0.07$ & $1.041\pm0.007$ \\
\hline
\multicolumn{6}{l}{\textbf{EMG}} \\
Top-3 distance
& $0.7645\pm0.0121$ & $0.2198\pm0.0097$
& $1.418\pm0.175$ & $84.32\pm2.25$ & $2.154\pm0.096$ \\
Top-3 uncertainty
& $0.7563\pm0.0006$ & $0.2061\pm0.0265$
& $1.344\pm0.254$ & $92.12\pm1.49$ & $2.092\pm0.222$ \\
\hline
\end{tabular}
\end{table}
Figure~\ref{fig:routing_distance_uncertainty_additional_metrics}
shows the corresponding forecasting error, likelihood
and interval-width comparisons.

\section{Additional selection-policy results}
\label{app:policy_results}

For reconstruction on incoming datasets, each encoder's intervals
use Gaussian moments from 40 posterior reconstruction draws.
We compute ECE and PICP90 separately for each selected encoder,
then average these metrics using the policy's aggregation weights,
as for reconstruction MSE in Equation~\ref{eq:policy_reconstruction_metric}.

\begin{figure}[tbp]
\centering
\begin{minipage}{0.80\linewidth}
    \centering
    \includegraphics[width=\linewidth]
    {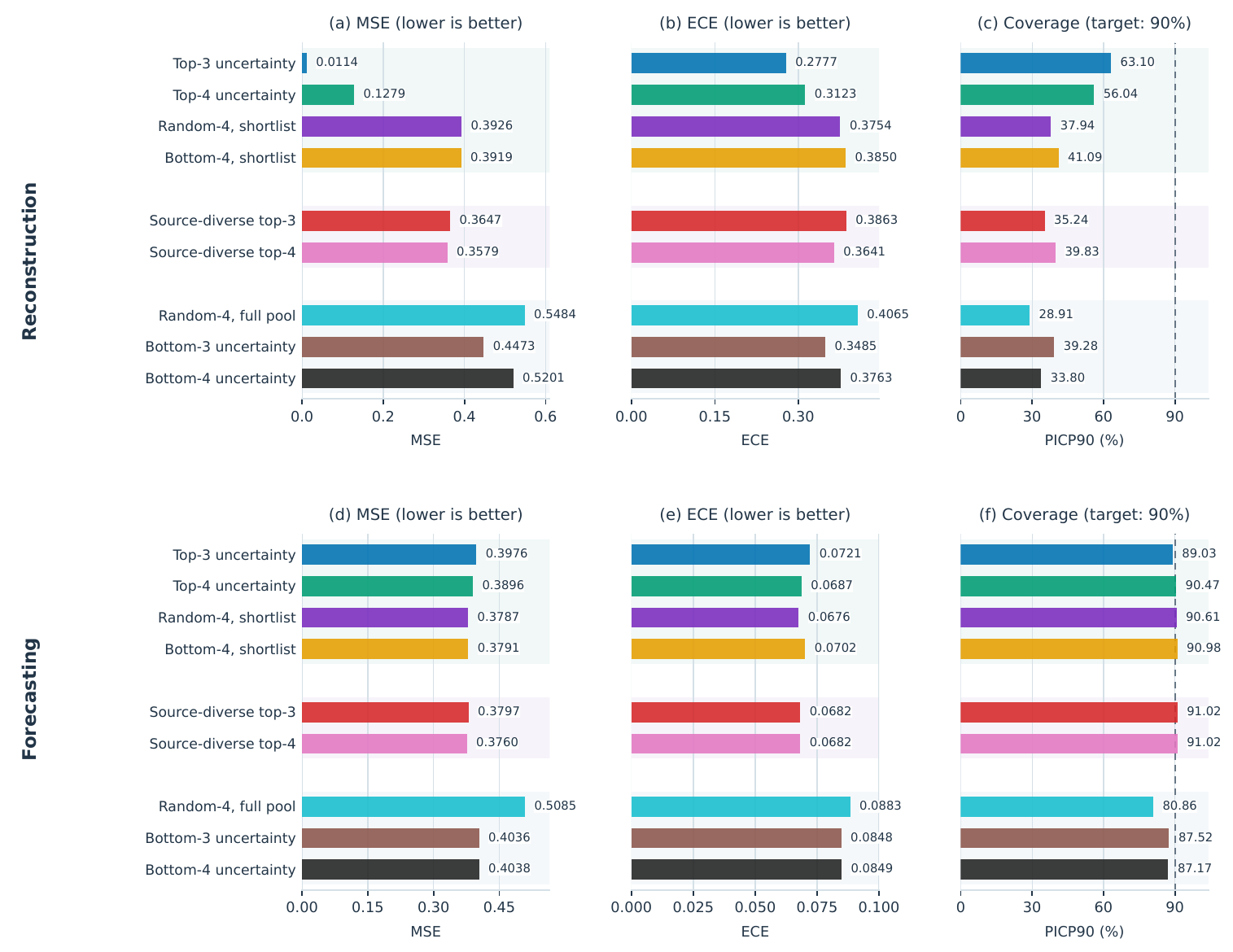}
\end{minipage}
\caption{\textbf{Reconstruction and forecasting under the selection policies.}
The upper row reports reconstruction on incoming datasets;
the lower row reports downstream forecasting.
Columns show MSE, ECE and PICP90.
Lower MSE and ECE are preferred; dashed lines mark nominal
90\% coverage. Policies retain the same colours and ordering.
}
\label{fig:policy_reconstruction_forecasting}
\end{figure}

\subsection{Detailed reconstruction and forecasting results}

Table~\ref{tab:policy_reconstruction_mse_by_dataset}
gives the per-dataset reconstruction MSE underlying
Table~\ref{tab:policy_reconstruction_results}.

\begin{table}[H]
\centering
\footnotesize
\setlength{\tabcolsep}{3pt}
\renewcommand{\arraystretch}{1.12}
\caption{\textbf{Reconstruction MSE on the incoming datasets.}
Bold marks the lowest error within each dataset.}
\label{tab:policy_reconstruction_mse_by_dataset}
\begin{tabular}{@{}lrrrrrrr@{}}
\hline
& \multicolumn{7}{c}{Reconstruction MSE $\downarrow$} \\
Policy & Torque10 & Torque90 & WTH & ETTh2 & ECL & Vibration & EMG \\
\hline
Top-3 uncertainty
& \textbf{0.0015} & $\mathbf{2.43\times10^{-6}}$
& \textbf{0.0268} & \textbf{0.0134}
& \textbf{0.0104} & \textbf{0.0184} & 0.0090 \\
Top-4 uncertainty
& 0.2069 & 0.1260 & 0.0832 & 0.1899 & 0.2602 & 0.0193 & 0.0098 \\
\hline
Source-diverse top-3
& 0.5247 & 0.3461 & 0.2405 & 0.5437 & 0.7143 & 0.0478 & 0.1360 \\
Source-diverse top-4
& 0.4926 & 0.3424 & 0.2060 & 0.4592 & 0.7136 & 0.1686 & 0.1226 \\
Random-4 of top-8
& 0.5989 & 0.4088 & 0.2092 & 0.6303 & 0.8713 & 0.0219 & \textbf{0.0079} \\
Bottom-4 of top-8
& 0.5975 & 0.4088 & 0.2068 & 0.6292 & 0.8713 & 0.0196 & 0.0098 \\
\hline
Random-4, full pool
& 0.7764 & 0.5433 & 0.2299 & 0.8573 & 1.1585 & 0.1625 & 0.1109 \\
Bottom-3 uncertainty
& 0.5527 & 0.3650 & 0.3177 & 0.6318 & 0.7781 & 0.4709 & 0.0152 \\
Bottom-4 uncertainty
& 0.6347 & 0.4084 & 0.3529 & 0.7104 & 0.8735 & 0.3966 & 0.2640 \\
\hline
\end{tabular}
\end{table}

Table~\ref{tab:policy_reconstruction_ece_by_dataset}
reports reconstruction ECE on the same incoming datasets.

\begin{table}[H]
\centering
\footnotesize
\setlength{\tabcolsep}{3pt}
\renewcommand{\arraystretch}{1.12}
\caption{\textbf{Calibration of incoming-data reconstructions.}
Bold marks the lowest ECE within each dataset.}
\label{tab:policy_reconstruction_ece_by_dataset}
\begin{tabular}{@{}lrrrrrrr@{}}
\hline
& \multicolumn{7}{c}{Reconstruction ECE $\downarrow$} \\
Policy & Torque10 & Torque90 & WTH & ETTh2 & ECL & Vibration & EMG \\
\hline
Top-3 uncertainty
& \textbf{0.3752} & \textbf{0.2316} & \textbf{0.1669}
& \textbf{0.1936} & \textbf{0.3404} & 0.2912 & 0.3451 \\
Top-4 uncertainty
& 0.4044 & 0.2933 & 0.2173 & 0.2596 & 0.3752 & 0.2931 & 0.3432 \\
\hline
Source-diverse top-3
& 0.4660 & 0.4250 & 0.3649 & 0.4045 & 0.4531 & 0.3129 & 0.2775 \\
Source-diverse top-4
& 0.4568 & 0.4237 & 0.3001 & 0.3459 & 0.4529 & 0.3129 & 0.2566 \\
Random-4 of top-8
& 0.4688 & 0.4179 & 0.3065 & 0.4082 & 0.4506 & 0.2996 & 0.2759 \\
Bottom-4 of top-8
& 0.4688 & 0.4179 & 0.3076 & 0.4081 & 0.4506 & 0.2943 & 0.3477 \\
\hline
Random-4, full pool
& 0.5000 & 0.5000 & 0.3079 & 0.4999 & 0.5000 & \textbf{0.2675} & 0.2700 \\
Bottom-3 uncertainty
& 0.4145 & 0.4128 & 0.3474 & 0.3466 & 0.4063 & 0.3701 & \textbf{0.1419} \\
Bottom-4 uncertainty
& 0.4359 & 0.4346 & 0.3829 & 0.3850 & 0.4298 & 0.3814 & 0.1848 \\
\hline
\end{tabular}
\end{table}

\paragraph{Forecasting calibration.}
Table~\ref{tab:policy_forecasting_ece_by_dataset} reports
forecasting ECE for each selection policy.

\begin{table}[H]
\centering
\footnotesize
\setlength{\tabcolsep}{3pt}
\renewcommand{\arraystretch}{1.15}
\caption{\textbf{Forecasting calibration by incoming dataset
and selection policy.}
Lower ECE is better. Bold values indicate the lowest ECE
within each dataset.}
\label{tab:policy_forecasting_ece_by_dataset}
\begin{tabular}{@{}lrrrrrrr@{}}
\hline
& \multicolumn{7}{c}{Forecasting ECE $\downarrow$} \\
Policy & Torque10 & Torque90 & WTH & ETTh2
& ECL & Vibration & EMG \\
\hline
Top-3 uncertainty
& 0.0419 & \textbf{0.0195} & 0.0055 & 0.0874
& \textbf{0.0067} & 0.1378 & \textbf{0.2061} \\

Top-4 uncertainty
& 0.0166 & 0.0327 & 0.0101 & 0.0643
& 0.0068 & 0.1345 & 0.2158 \\

Source-diverse top-3
& \textbf{0.0095} & 0.0326 & 0.0161 & 0.0485
& 0.0135 & 0.1273 & 0.2296 \\

Source-diverse top-4
& 0.0129 & 0.0401 & 0.0108 & 0.0514
& 0.0098 & 0.1240 & 0.2285 \\

Random-4 of top-8
& 0.0118 & 0.0345 & \textbf{0.0052} & 0.0593
& 0.0186 & 0.1171 & 0.2263 \\

Bottom-4 of top-8
& 0.0181 & 0.0347 & 0.0081 & 0.0516
& 0.0132 & \textbf{0.1146} & 0.2509 \\

Random-4, full pool
& 0.0974 & 0.0560 & 0.0094 & 0.0426
& 0.0386 & 0.1298 & 0.2447 \\

Bottom-3 uncertainty
& 0.0382 & 0.0695 & 0.0229 & \textbf{0.0380}
& 0.0223 & 0.1368 & 0.2656 \\

Bottom-4 uncertainty
& 0.0357 & 0.0766 & 0.0140 & 0.0552
& 0.0233 & 0.1244 & 0.2649 \\
\hline
\end{tabular}
\end{table}

\subsection{Routing with a fixed uncertainty-ratio cutoff}
\label{app:ratio_threshold_routing}

We also evaluate the policy \texttt{ratio\_le\_1p10}.
From the pool of 17 encoders, it selects every encoder
whose source-normalized uncertainty ratio satisfies
$r_m\leq1.10$.
Selected encoders receive normalized inverse-ratio weights,
following Equation~\ref{eq:inverse_ratio_routing_weights}.

The cutoff was fixed from reconstruction-stage uncertainty
ratios before forecasting.
This policy allows the number of selected encoders to vary
across incoming datasets.
It is separate from the eight-member shortlist used
in the nine-policy comparison.
The forecasting evaluation uses seeds 40, 41 and 42. Table~\ref{tab:ratio_threshold_forecasting} compares the cutoff policy with fixed top-$3$ and top-$4$ uncertainty routing.  \begin{table}[H] \centering \small \setlength{\tabcolsep}{3pt} \renewcommand{\arraystretch}{1.12} \caption{\textbf{Fixed-$K$ and fixed-cutoff uncertainty routing.} The cutoff policy selects all encoders with $r_m\leq1.10$. Entries are means and sample standard deviations across the three seed-specific averages over seven incoming datasets.} \label{tab:ratio_threshold_forecasting} \begin{tabular}{@{}lccccc@{}} \hline Policy & MSE $\downarrow$ & ECE $\downarrow$ & \shortstack{PICP90 (\%)\\$\to90\%$} & NLL $\downarrow$ & MPIW90 \\ \hline Top-$3$ & $0.3976\pm0.0070$ & $0.0721\pm0.0038$ & $89.03\pm0.35$ & $0.8357\pm0.0413$ & $1.715\pm0.032$ \\ Top-$4$ & $0.3896\pm0.0031$ & $0.0687\pm0.0025$ & $90.47\pm0.09$ & $0.8135\pm0.0213$ & $1.771\pm0.018$ \\ Cutoff $1.10$ & $0.3972\pm0.0070$ & $0.0719\pm0.0045$ & $89.03\pm0.23$ & $0.8240\pm0.0125$ & $1.716\pm0.020$ \\ \hline \end{tabular} \end{table}

\begin{table}[H]
\centering
\small
\renewcommand{\arraystretch}{1.12}
\caption{\textbf{Expert selection with a fixed uncertainty cutoff.}
The eligible pool contains 17 encoders.
The policy retains encoders with $r_m\leq1.10$.}
\label{tab:ratio_threshold_selection}
\begin{tabular}{@{}lr@{}}
\hline
Incoming dataset & Selected encoders \\
\hline
Torque10  & 3 \\
Torque90  & 3 \\
WTH       & 6 \\
ETTh2     & 3 \\
ECL       & 3 \\
Vibration & 10 \\
EMG       & 10 \\
\hline
\end{tabular}
\end{table}
\end{document}